\documentclass{article} % For LaTeX2e
\usepackage{iclr2026_conference,times}
\iclrfinalcopy

\usepackage{amsmath,amsfonts,bm}

\def\eqref#1{equation~\ref{#1}}
\def\1{\bm{1}}

\DeclareMathAlphabet{\mathsfit}{\encodingdefault}{\sfdefault}{m}{sl}
\SetMathAlphabet{\mathsfit}{bold}{\encodingdefault}{\sfdefault}{bx}{n}

\PassOptionsToPackage{numbers, compress}{natbib}

\usepackage[utf8]{inputenc} % allow utf-8 input
\usepackage[T1]{fontenc}    % use 8-bit T1 fonts
\usepackage{booktabs}       % professional-quality tables
\usepackage{amsfonts}       % blackboard math symbols
\usepackage{nicefrac}       % compact symbols for 1/2, etc.
\usepackage{microtype}      % microtypography
\usepackage{times}
\usepackage{CJKutf8}
\usepackage{latexsym}
\usepackage[most]{tcolorbox}
\usepackage{wrapfig}
\usepackage{capt-of}
\usepackage{pgfplots}
\pgfplotsset{compat=1.12}
\usepackage{amsmath}
\usepackage{multicol}
\usepackage{color}
\usepackage{mwe}
\usepackage{wrapfig}
\usepackage{colortbl,array}
\usepackage{xspace}
\usepackage{hyperref}
\usepackage{url}
\usepackage{xcolor}
\usepackage{enumitem}
\usepackage{graphicx}
\usepackage{booktabs}
\usepackage{multirow}
\usepackage{fontawesome}
\usepackage[table]{xcolor}
\usepackage{float}
\usepackage{tikz}
\usepackage{algorithm}
\usepackage{algorithmicx}
\usepackage{algpseudocode}

\usetikzlibrary{tikzmark}

\definecolor{teaserred}{RGB}{180,10,56}

\definecolor{teaserblue}{RGB}{0,15,139}

\definecolor{uclablue}{RGB}{159, 195, 224}

\definecolor{uclagold}{RGB}{254,180,167}

\definecolor{grayred}{RGB}{232,237,205}

\definecolor{qing}{RGB}{176, 227, 230}

\definecolor{TealBlue}{rgb}{1.0, 0.97, 0.8}

\makeatletter
\newcommand*\myfontsize{%
  \@setfontsize\myfontsize{7}{8}%
}
\makeatother

\newcommand{\mytextbox}[2]{%
  \tikzmarknode[draw=#1,thick,inner sep=2pt]{mybox}{\myfontsize #2}%
}
\definecolor{mypurple}{RGB}{153, 51, 255}
\definecolor{myorange}{RGB}{255, 153, 51}
\definecolor{mygreen}{RGB}{0, 153, 0}
\definecolor{myblue}{RGB}{108, 142, 191}
\definecolor{mygreen2}{RGB}{60, 179, 113}

\newcommand{\purple}[1]{\mytextbox{mypurple}{\textbf{\textcolor{mypurple}{#1}}}}
\newcommand{\orange}[1]{\mytextbox{myorange}{\textbf{\textcolor{myorange}{#1}}}}
\newcommand{\green}[1]{\mytextbox{mygreen}{\textbf{\textcolor{mygreen}{#1}}}}
\newtcolorbox[auto counter, number within=section]{promptbox}[2][]{%
  colback=white, 
  colframe=myblue,  
  width=\textwidth,
  arc=2mm, 
  boxrule=0.5mm, 
  title={\normalsize\faInfoCircle\hspace{0.5em}#2},
  breakable,
  fonttitle=\bfseries\Large, 
  fontupper=\small, 
  drop shadow southeast, 
  #1
}
\newtcolorbox[auto counter, number within=section]{purposebox}[2][]{%
  colback=white, 
  colframe=mygreen2,  
  width=\textwidth,
  arc=2mm, 
  boxrule=0.5mm, 
  title={\normalsize\faInfoCircle\hspace{0.5em}#2},
  breakable,
  fonttitle=\bfseries\Large, 
  fontupper=\small, 
  drop shadow southeast, 
  #1
}
\newcommand\best[1]{\textcolor{red}{\textbf{#1}}}
\newcommand\secbest[1]{\textcolor{blue}{\underline{#1}}}

\title{Video-Thinker: Sparking ``Thinking with Videos'' via Reinforcement Learning}

\author{Shijian Wang$^{1,2,3}$\thanks{Equal contribution. Work done when Shijian internship at Xiaohongshu Inc.}, Jiarui Jin$^{3*}$, Xingjian Wang$^{2}$, Linxin Song$^{4}$, Runhao Fu$^{2}$, \\
\textbf{Hecheng Wang}$^{5}$, \textbf{Zongyuan Ge}$^{2}$, \textbf{Yuan Lu}$^{3}$\thanks{Corresponding Authors}, \textbf{Xuelian Cheng}$^{2\dagger}$ \\
$^{1}$Southeast University, $^{2}$Monash University, $^{3}$Xiaohongshu Inc., \\
$^{4}$University of Southern California, $^{5}$Fudan University\\
\small{\texttt{\{wangshijian,jinjiarui,luyuan3\}@xiaohongshu.com}}\\[1mm]\\
\includegraphics[height=4mm]{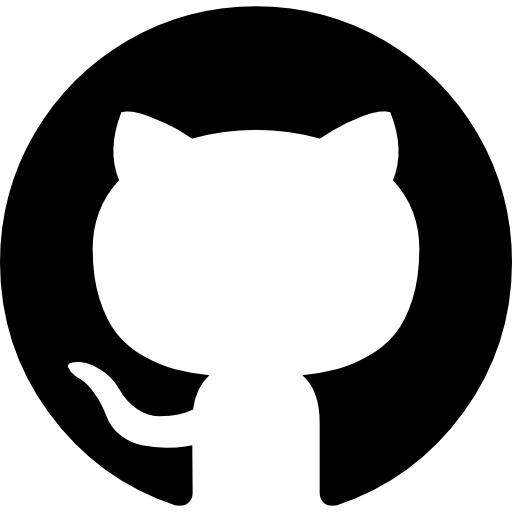} Code: \href{https://github.com/shijian2001/Video-Thinker}{\small{\texttt{shijian2001/Video-Thinker}}}
\raisebox{-1mm}{\includegraphics[height=5mm]{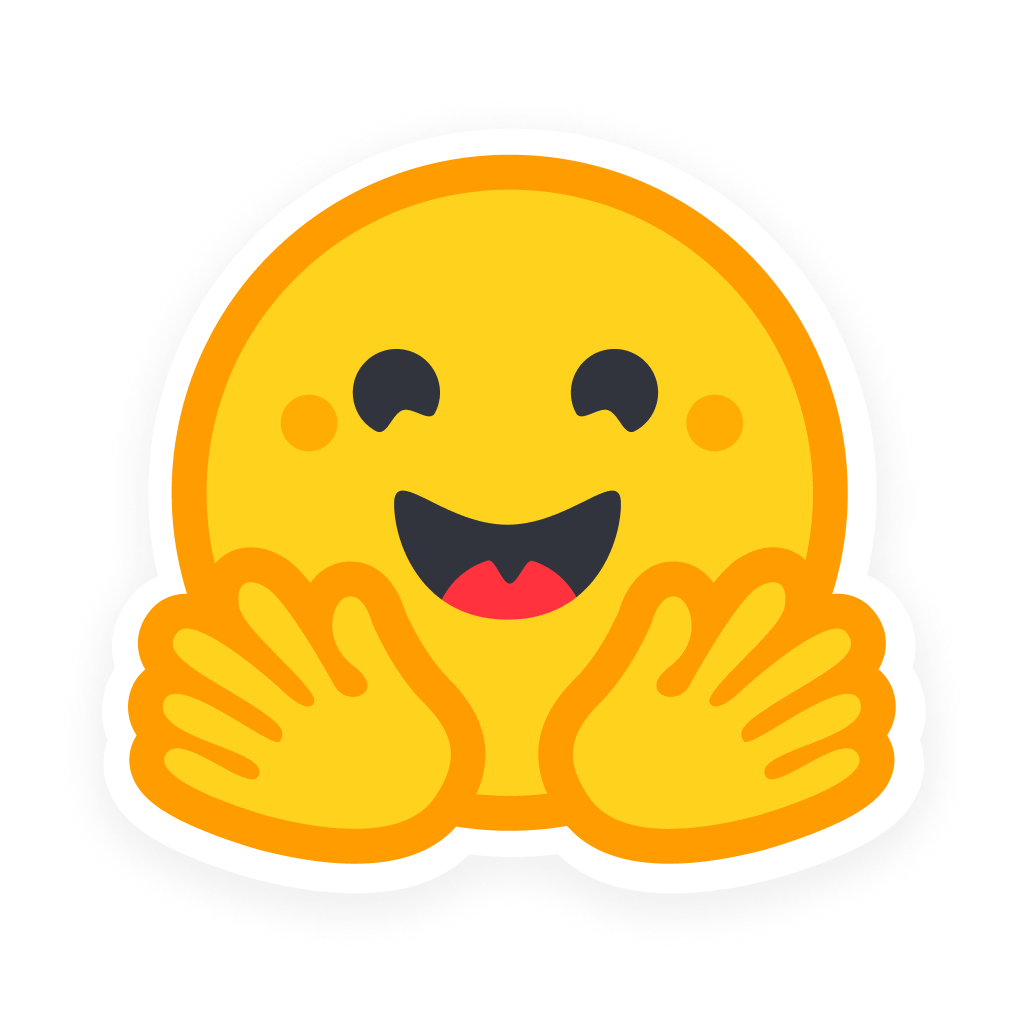}} Model: \href{https://huggingface.co/ShijianW01/Video-Thinker-7B}{\small{\texttt{ShijianW01/Video-Thinker-7B}}}
}

\newcommand{\highlight}[1]{\noindent\textbf{#1}}

\begin{document}

\maketitle

\begin{abstract}
Recent advances in image reasoning methods, particularly ``Thinking with Images'', have demonstrated remarkable success in Multimodal Large Language Models (MLLMs); however, this dynamic reasoning paradigm has not yet been extended to video reasoning tasks. In this paper, we propose Video-Thinker, which empowers MLLMs to think with videos by autonomously leveraging their intrinsic ``grounding'' and ``captioning'' capabilities to generate reasoning clues throughout the inference process. To spark this capability, we construct Video-Thinker-10K, a curated dataset featuring autonomous tool usage within chain-of-thought reasoning sequences. Our training strategy begins with Supervised Fine-Tuning (SFT) to learn the reasoning format, followed by Group Relative Policy Optimization (GRPO) to strengthen this reasoning capability. Through this approach, Video-Thinker enables MLLMs to autonomously navigate grounding and captioning tasks for video reasoning, eliminating the need for constructing and calling external tools. Extensive experiments demonstrate that Video-Thinker achieves significant performance gains on both in-domain tasks and challenging out-of-domain video reasoning benchmarks, including Video-Holmes, CG-Bench-Reasoning, and VRBench. Our Video-Thinker-7B substantially outperforms existing baselines such as Video-R1 and establishes state-of-the-art performance among 7B-sized MLLMs.
\end{abstract}
% \begin{abstract}
% Recent advances have enabled large Multimodal Language Models (MLMs) to ``Think with Images" for image reasoning, yet this dynamic reasoning paradigm does not extend to videos, where MLMs' inability to manipulate temporal information forms a critical bottleneck. To address this issue, we introduce \textbf{\textit{Video-Thinker}}, a framework that empowers MLMs to ``Think with Videos" by endowing them with intrinsic capabilities for \textit{time grounding} and \textit{ video captioning} within a chain-of-thought thinking process. To foster these skills, we construct \textbf{\textit{Video-Thinker-10K}}, a curated dataset featuring structured reasoning traces. Our training strategy begins with Supervised Fine-Tuning (SFT) to learn the reasoning format, followed by using Group Relative Policy Optimization (GRPO) to scale this temporal reasoning paradigm, where only the final answer serves as the outcome reward. Our approach compels MLM to autonomously navigate temporal localization and summarization capabilities for video reasoning, eliminating the need for external tools. Extensive experiments show that \textbf{\textit{Video-Thinker}} achieves significant gains on both in-domain tasks and challenging out-of-domain benchmarks such as Video-Holmes, MMVU, VideoMMMU, LongVideoBench and VSI-Bench. Notably, our 7B model achieves X\% on Video-Holmes, outperforming the strong Video-R1 baseline.
% \end{abstract}

\section{Introduction}
\label{sec:intro}

\begin{wrapfigure}{r}{0.58\textwidth}
    \centering
    \includegraphics[width=0.55\textwidth,page=1]{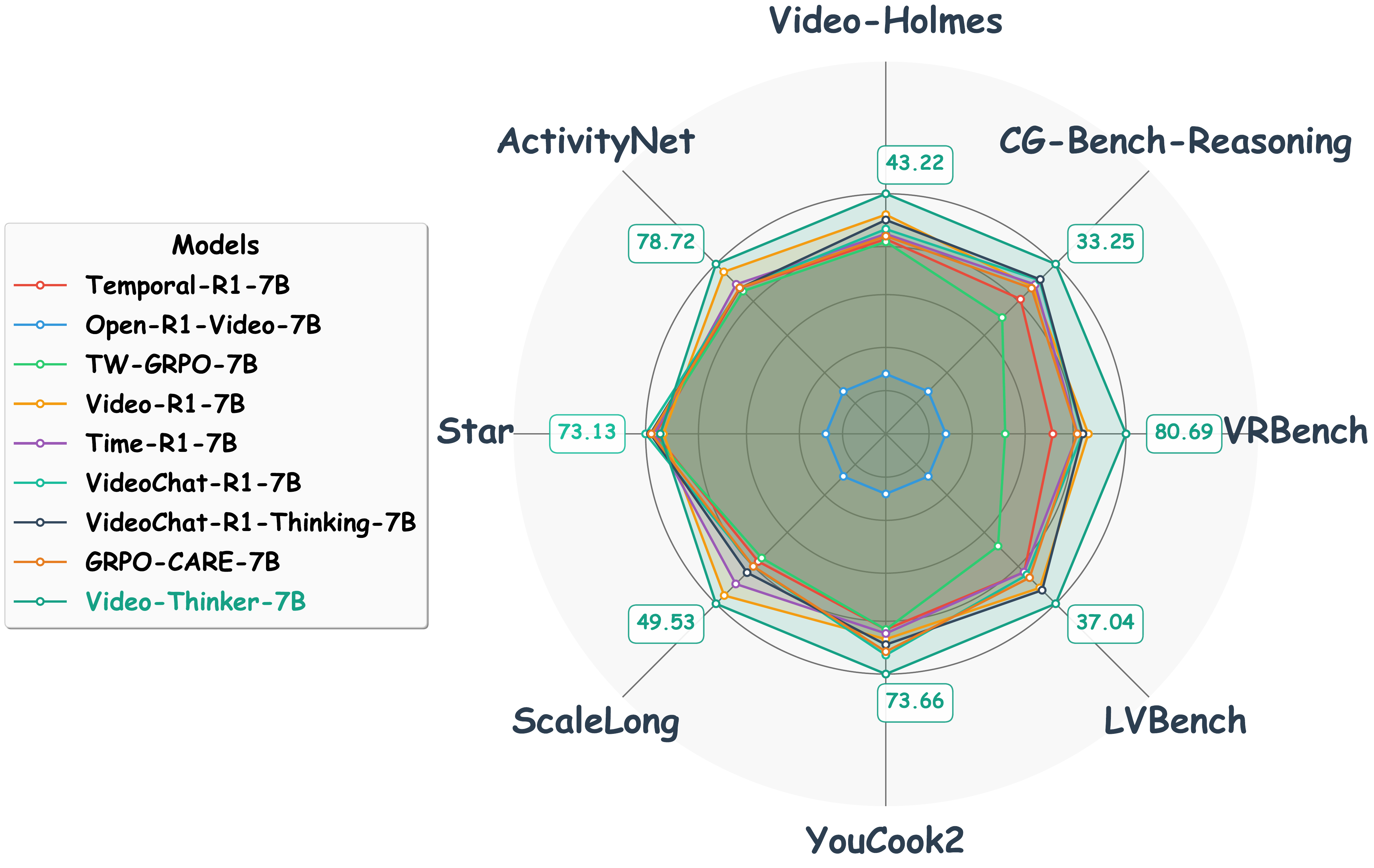}
    \caption{Overall Performance of Video-Thinker}
    \label{fig:performance}
\end{wrapfigure}

Multimodal Large Language Models (MLLMs) have embraced a revolutionary paradigm shift toward ``Thinking with Images'' for image understanding and reasoning tasks, evolving from passively treating images as static context to actively localizing, zooming in, and reasoning over image content during the thinking process~\citep{deepeyes, llavaplus, zoomeye, visuothink, taco}. 
This dynamic multimodal reasoning paradigm has yielded substantial advances on MLLMs across diverse image reasoning tasks, including visual question answering~\citep{llava, cotvla, visualprog, ii-bench}, visual mathematical problem solving~\citep{mint, visualcot, mathcoder, mmmu, visiomath, an2025unictokens}, and complex scene understanding~\citep{luo2024llm, ferret, som, puzzlebench, deepeyes, iv, lin2025perceive}. 
However, the extension of these capabilities to video understanding presents significant challenges.
Unlike static images, videos inherently contain temporal dependencies, motion patterns, and evolving visual narratives that require sophisticated temporal reasoning mechanisms, whereas MLLMs struggle to dynamically manipulate and reason over temporal sequences without relying on explicitly pre-designed chain-of-thought prompting strategies \citep{fei2024video,video-r1,shi2024enhancing, an2024mc}.
% This limitation in effectively leveraging temporal dynamics represents a critical bottleneck in advancing video reasoning capabilities.

\begin{figure*}[t]
    \vspace{-9mm}
    \centering
    \includegraphics[width=\textwidth]{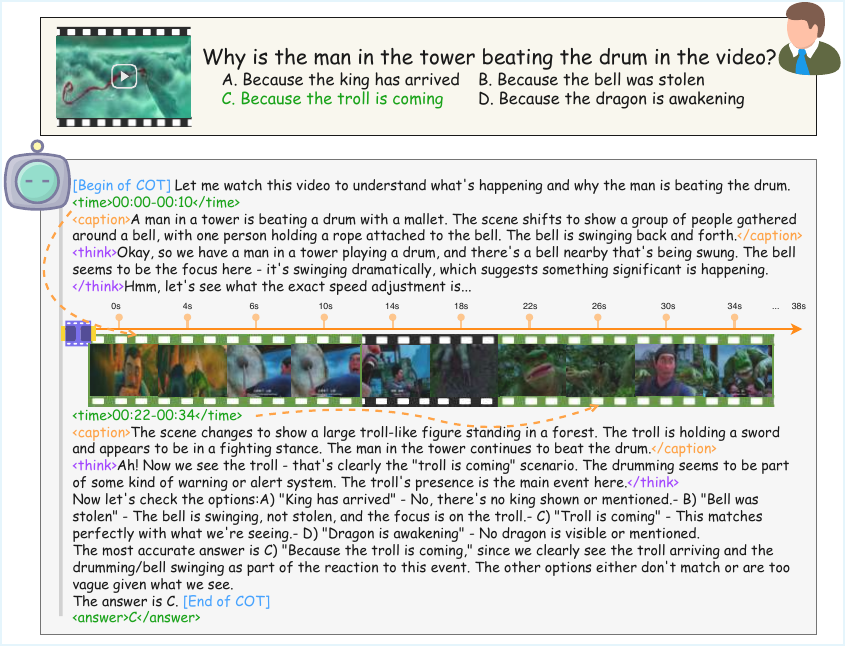}
    \vspace{-6mm}
    \caption{Video-Thinker integrates ``grounding'' and ``captioning'' capabilities throughout the reasoning process using end-to-end reinforcement learning.}
    \label{fig:wide-pdf}
    \vspace{-4mm}
\end{figure*}

In this paper, we propose a novel framework named \textbf{Video-Thinker} to enhance MLLMs by enabling them to perform visual reasoning through structured video analysis capabilities.
Drawing inspiration from spatial visual operations in  ``Thinking with Images''~\citep{openaio3} for image understanding — such as ``crop'' for region localization and ``zoom-in'' for detailed region comprehension — we introduce the following temporal visual operations - namely ``grounding'' and ``captioning''.
The ``grounding'' operation serves as a temporal localization mechanism that identifies and extracts key frames containing critical visual information within the video sequence, while the ``captioning'' operation functions as a comprehension mechanism that analyzes these key frames to extract, interpret, and synthesize relevant visual cues into a coherent understanding.
Fortunately, these video localization and comprehension capabilities can be developed within MLLMs themselves, thereby eliminating the need for MLLMs to adapt to and invoke external handcrafted tools.
Hence, our Video-Thinker can enable structured temporal reasoning through chain-of-thought (CoT) processes, allowing models to autonomously navigate and analyze specific temporal segments rather than treating videos as monolithic inputs. 
The framework orchestrates these temporal manipulation capabilities through systematic reasoning traces that synthesize visual cues across multiple video segments.
Our approach differs fundamentally from previous investigations in two key aspects. 
First, unlike video-of-thoughts methodologies that rely on sophisticated pre-designed CoT processes \citep{fei2024video}, our framework develops intrinsic temporal reasoning capabilities that emerge naturally from the training process. 
Second, in contrast to general visual reasoning models that require extensive datasets exceeding 160K samples \citep{video-r1}, our approach demonstrates that effective video reasoning capabilities can be achieved with significantly greater efficiency using only 10K carefully curated training examples.

% Recent breakthroughs on ``Thinking with Images" exemplified by OpenAI's O3 demonstrate that integrating visual actions directly into the thinking process yields superior performance in complex image reasoning tasks. 
% O3's success stems from its effective use of visual tools such as region localization for object identification and zoom-in operations for detailed examination. 
% Analogously, video reasoning demands corresponding temporal capabilities: \textit{time grounding} for temporal segment localization and \textit{video captioning} for temporal content summarization. 

% Our framework, which integrates time grounding and video captioning as core competencies within the MLLM architecture, empowers MLLMs with robust video reasoning capabilities while maintaining computational and data efficiency.
To instantiate our framework, we carefully construct \textbf{Video-Thinker-10K}, a curated training dataset of 10K samples spanning diverse video-reasoning tasks and domains. 
Each sample comprises strategically selected key video segments, detailed captions describing visual clues for each temporal window, and structured reasoning traces that demonstrate how to synthesize these multimodal cues for complex video understanding tasks.
As illustrated in Figure~\ref{fig:wide-pdf}, our reasoning trace adopts a structured format wherein each key video segment is systematically processed through three specialized annotation tags: the \texttt{\green{<time></time>}} tag for precise temporal localization, the \texttt{\orange{<caption></caption>}} tag for comprehensive visual cue extraction, and the \texttt{\purple{<think></think>}} tag for analytical reasoning that synthesizes the extracted visual information.

Our training methodology employs a two-stage approach: we first conduct supervised fine-tuning (SFT) using our curated thought processes as ground truth supervision to establish foundational format-following capabilities. 
We subsequently apply Group Relative Policy Optimization (GRPO)~\citep{deepseekmath} for reinforcement learning, where only the final answer serves as the outcome reward. 
This approach enables the model to intrinsically acquire both grounding and captioning capabilities, facilitating autonomous temporal navigation for sophisticated video reasoning tasks.
Our extensive experiments demonstrate that Video-Thinker achieves the state-of-the-art (SOTA) performance among 7B-sized MLLMs across various challenging out-of-domain video reasoning benchmarks, including Video-Holmes~\citep{Video-Holmes}, CG-Bench-Reasoning~\cite{cgbench}, and VRBench~\citep{vrbench}, as demonstrated in Figure~\ref{fig:performance}. 
% Furthermore, we observe that throughout the training process, our model's intrinsic \textit{time grounding} and \textit{video captioning} capabilities progressively enhance, which validates that our framework effectively integrates these temporal operations into the model's inherent reasoning capabilities, enabling autonomous video reasoning without relying on external tools.

Our main contributions are summarized as follows: (i) proposing a new paradigm (Video-Thinker) of ``Thinking with Videos'' by intrinsically integrating grounding and captioning capabilities within the CoT process, eliminating the dependency on external tools; (ii) contributing a meticulously curated video reasoning dataset (Video-Thinker-10K) encompassing comprehensive localization annotations and rich comprehension information; and (iii) empirically setting new SOTA performances across multiple video reasoning benchmarks.

\section{Related Work}
\label{sec:related}
Recent advances in reinforcement learning-based post-training have demonstrated significant improvements in reasoning capabilities, as evidenced by OpenAI-o1 \citep{jaech2024openai} and Deepseek-R1 \citep{guo2025deepseek}. 
Building upon this foundation, the field of MLLMs is undergoing a paradigmatic shift in how visual information is integrated into reasoning processes.
Traditionally, MLLMs have treated images as static inputs, relegating the reasoning process entirely to the textual domain \citep{thinking}. 
An emerging paradigm, however, elevates visual information to an explicit, manipulable intermediate within the reasoning process itself, transforming vision from a passive input into an active cognitive tool \citep{openaio3}. 
This approach is exemplified by several recent works: Deepeyes \citep{deepeyes} employs end-to-end reinforcement learning to train models that autonomously invoke visual tools (e.g., magnification) while interleaving visual and textual CoT reasoning, effectively enabling models to ``Think with Images''. 
Visual-ARFT \citep{visualarft} utilizes GRPO \citep{deepseekmath} to develop capabilities in task planning, stepwise reasoning, and tool use, allowing models to strategically employ Python-based image-processing operators. 

The natural extension of these advances lies in video reasoning, which represents a core capability for MLLMs seeking to capture the logical structure of temporal visual content—a crucial step beyond mere video perception toward genuine video understanding \citep{open-r1-video, tw-grpo, videoutr}. 
Recent efforts have begun addressing this challenge: 
Video-R1 \citep{video-r1} extends GRPO into the video domain, promoting implicit temporal reasoning alongside spatial reasoning capabilities. 
VideoChat-R1 \citep{videochatr1} leverages reinforcement fine-tuning to strengthen spatiotemporal localization while preserving conversational proficiency. 
Temporal-R1 \citep{temporal} employs explicit temporal grounding rewards and variance-aware data selection strategies to enhance both semantic and temporal reasoning with improved data efficiency.

Despite these advances, current approaches remain largely confined to either temporal localization or standalone video reasoning, falling short of integrating temporal grounding seamlessly into the CoT processes. 
Our proposed Video-Thinker framework — extending the paradigm of ``Think with Images'' — enables MLLMs to ``Think with Videos'' by facilitating dynamic navigation of temporal content within the reasoning process.
Specifically, Video-Thinker incorporates ``grounding'' and ``captioning'' capabilities as integral components of the CoT reasoning, allowing MLLMs to systematically attend to, interpret, and analyze relevant temporal segments throughout video-based tasks.

\section{Think with Videos: From Data Synthesis to Model Training}
\label{sec:method}
As video reasoning tasks require temporal localization and comprehension capabilities in MLLMs, we propose ``grounding'' and ``captioning'' as fundamental anchors for model enhancement. 
To address this requirement, we first establish high-quality curated data termed Video-Thinker-10K, using a new hindsight-curation reasoning method, as detailed in Section~\ref{sec:dataset}. 
Subsequently, we train our Video-Thinker models on these datasets through supervised fine-tuning and reinforcement learning approaches, as described in Section~\ref{sec:training}.

\subsection{Data Synthesis via Hindsight-curation Reasoning}
\label{sec:dataset}
Here, we curate a diverse collection of source data from the following six prominent datasets, namely ActivityNet~\citep{caba2015activitynet}, TutorialVQA~\citep{colas2019tutorialvqa}, YouCook2~\citep{zhou2018towards}, STAR~\citep{wu2024star}, ScaleLong~\citep{ma2025scalelong}, and LVBench~\citep{wang2024lvbench}. 
These sources span a wide spectrum of domains — ranging from human activities and instructional tutorials to cooking procedures, situated reasoning, and long-form content such as TV series. 
Within these datasets, we identified the following two complementary categories of data:
(i) Caption-labeled datasets, including ActivityNet, TutorialVQA, and YouCook2, provide detailed, human-annotated captions for specific temporal intervals within key video segments but lack complex questions that require deep reasoning capabilities.
(ii) QA-labeled datasets, comprising STAR, ScaleLong, and LVBench, offer challenging question-answer pairs designed for deep reasoning but lack the granular, per-segment visual descriptions essential for our structured reasoning framework.

\begin{figure*}[t]
    \vspace{-9mm}
    \centering
    \includegraphics[width=\textwidth]{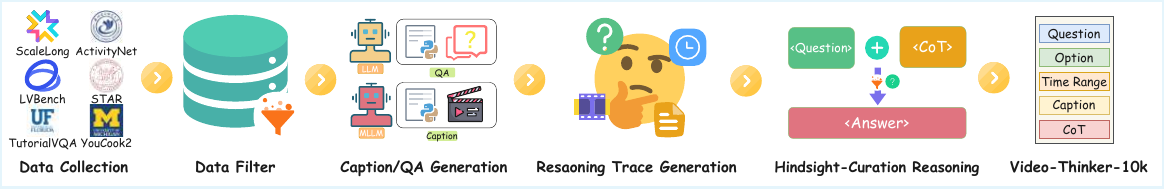}
    \vspace{-5mm}
    \caption{Data synthesis pipeline of Video-Thinker-10K where the data distribution is depicted in Figure~\ref{fig:dataset} in Appendix~\ref{app:dataset}.}
    \label{fig:data-pipe}
    \vspace{-4mm}
\end{figure*}

To inspire MLLMs with intrinsic capabilities for ``grounding'' and ``captioning'', our training data curation is guided by two core principles. 
One is: our training data requires questions that compel MLLMs to localize multiple key segments, accurately summarize their content, and synthesize this information to derive comprehensive answers. 
The other one is: our training data must provide supervision through a structured reasoning trace that includes the \texttt{\green{<time></time>}} tag for temporal localization, the \texttt{\orange{<caption></caption>}} tag for visual cue description, and the \texttt{\purple{<think></think>}} tag for analytical reasoning, explicitly integrating temporal actions within the CoT process.
To bridge the gap between the collected source data and the expected structured data samples described above, we developed a systematic data transformation pipeline, as demonstrated in Figure~\ref{fig:data-pipe}).

We first applied quality filters to remove corrupted videos and exclude videos with fewer than 64 frames to ensure adequate temporal content. Our pipeline then branches into two distinct generation strategies based on dataset characteristics:
(i) For caption-labeled datasets (namely, ActivityNet, TutorialVQA, YouCook2) that are rich in temporal annotations and segment descriptions, we focused on synthesizing corresponding reasoning questions. 
We leveraged DeepSeek-R1~\citep{deepseekr1} to generate complex multiple-choice questions that necessitate reasoning across multiple video segments, using the existing detailed segment descriptions as the contextual foundation.
(ii) For QA-labeled datasets (namely, STAR, ScaleLong, LVBench) that provide high-quality question-answer pairs but lack granular per-segment descriptions, we concentrated on generating the missing visual cues. 
Given the ground-truth answers and temporal annotations, we employed Gemini-2.5-Flash-Lite~\citep{comanici2025gemini} to produce answer-conditioned descriptive captions for video segments, ensuring that the generated visual descriptions are relevant to the reasoning process.

Finally, with both question-answer pairs and segment-level visual descriptions now available across all data samples, we perform the final reasoning trace synthesis. 
We use DeepSeek-V3~\citep{liu2024deepseek} for reverse-curation generation, where the model receives the ground-truth answer, generated visual descriptions (captions), and temporal annotations to produce high-quality reasoning processes that articulate step-by-step temporal analysis. 
Each trace adheres to our predefined structured format, incorporating the \texttt{\green{<time></time>}} tag for temporal localization, the \texttt{\orange{<caption></caption>}} tag for visual evidence summarization, and the \texttt{\purple{<think></think>}} tag for analytical reasoning elaboration, thereby creating complete training instances for our Video-Thinker-10K dataset. 

To ensure that the generated ``grounding'' and ``captioning'' components are beneficial for the final response, previous data synthesis pipelines such as Video-Holmes~\citep{Video-Holmes} employ manual sampling inspection to ensure quality and relevance. 
To reduce the cost of human evaluation and annotation, we propose a novel hindsight curation process. For each sample, the generated content within the \texttt{\green{<time></time>}} and \texttt{\orange{<caption></caption>}} tags is input into Qwen2.5-VL-7B-Instruct~\citep{Qwen2.5-VL} to evaluate whether the model can derive the correct answer. 
If the model fails to produce the accurate answer, we regenerate the reasoning trace.
This iterative process repeats up to three times, ensuring that all samples are equipped with a high-quality and relevant reasoning trace that effectively guides the model toward the correct solution.
Also, we carefully sample from these sources to ensure a balanced distribution across various tasks and domains, as detailed in Figure~\ref{fig:dataset} in Appendix~\ref{app:dataset}.
We also provide the specific prompt templates used in this generation pipeline in Appendix~\ref{app:prompt}.

\subsection{Training Strategy of Video-Thinker}
\label{sec:training}
Let $D=(V,Q,T,Y)\in\mathcal{D}_\text{Video-Thinker}$ denote any sample in Video-Thinker-10K constructed in the above subsection, where $V$ represents the video, $Q$ is the question, $T$ is the ground-truth reasoning trace containing grounding and captioning contents, and $Y$ is the ground-truth answer.  

\highlight{SFT Optimization for Format-Following.}
We start by Supervised Fine-tuning (SFT) to bootstrap Video-Thinker's ability to generate structured reasoning traces over ``grounding'' and ``captioning'' contents. 
Since pre-trained MLLMs lack exposure to our specialized reasoning format with \texttt{\green{<time></time>}}, \texttt{\orange{<caption></caption>}}, and \texttt{\purple{<think></think>}} tags, SFT provides essential cold-start initialization by teaching the model to follow high-quality reasoning patterns from our Video-Thinker-10K dataset.

Formally, the SFT objective is to minimize the negative log-likelihood of the target reasoning trace $T$ and final answer $Y$, where the loss function can be formulated as:
\begin{equation}
\label{eq:sft}
\mathcal{L}_{\text{SFT}}(\theta) = - \mathbb{E}_{(V, Q, Y) \sim \mathcal{D}_\text{Video-Thinker}} \left[ \sum_{t=1}^{|[T;Y]|} \log p_{\theta}\Bigg([T;Y]_t \Bigg| V, Q, [T;Y]_{<t}\Bigg) \right],
\end{equation}
where $[T;Y]$ denotes the concatenation of $T$ and $Y$, and $p_{\theta}$ is the policy of Video-Thinker model parameterized by $\theta$. 
Namely, the model is trained to predict each subsequent token $[T;Y]_t$ of the reasoning trace and the final answer, conditioned on the video $V$, the question $Q$, and the preceding tokens $[T;Y]_{<t}$.

\highlight{GRPO Optimization for Autonomous Navigation over Grounding and Captioning Capabilities.}
To achieve sophisticated video reasoning with autonomous navigation over grounding and captioning capabilities, we employ Group Relative Policy Optimization (GRPO) to further optimize Video-Thinker beyond the above SFT stage. 
GRPO eliminates the need for value function approximation by generating multiple candidate responses for each $(V, Q, Y)$ sample and assessing their relative quality through verifiable rewards.
Formally, for each $(V, Q, Y)$ sampled from $\mathcal{D}_\text{Video-Thinker}$, GRPO generates $G$ distinct reasoning traces $\{T^{(1)}, T^{(2)}, \ldots, T^{(G)}\}$ using the current policy $p_{\theta_{\text{old}}}$. The policy is optimized by maximizing:

\begin{equation}
\label{eqn:grpo}
\begin{aligned}
    \mathcal{J}_{\text{GRPO}}(\theta) = 
    \mathbb{E}_{(V, Q, T,Y) \sim \mathcal{D}_\text{Video-Thinker}} \Bigg[ \frac{1}{G} \sum_{i=1}^G \Bigg( 
    & \min \Big( \frac{\pi_\theta}{\pi_{\theta_{\text{old}}}} A_i,\quad \text{clip}\Big( \frac{\pi_\theta}{\pi_{\theta_{\text{old}}}} , 1 - \epsilon, 1 + \epsilon \Big) A_i \Big) \\
    & \quad - \beta \, \text{KL}\Big(p_\theta(\cdot | V, Q) \Big\| p_{\text{ref}}(\cdot | V, Q)\Big) \Bigg) \Bigg],
\end{aligned}
\end{equation}
where $\pi_\theta = p_\theta(T^{(i)} | V, Q)$, $\pi_{\theta_{\text{old}}} = p_{\theta_{\text{old}}}(T^{(i)} | V, Q)$, $\text{KL}(p_\theta(\cdot | V, Q) \| p_{\text{ref}}(\cdot | V, Q))$ denotes the KL divergence \citep{van2014renyi} between the current policy $p_\theta(\cdot | V, Q)$ and reference policy $p_{\text{ref}}(\cdot | V, Q))$, $A_i$ is the advantage for the $i$-th reasoning trace, and $\epsilon$ and $\beta$ are hyperparameters
Here, the advantage $A_i$ is computed using outcome supervision based on normalized rewards within each group. 
Specifically, for each reasoning trace $T^{(i)}$, we assign a reward $r^{(i)}$ comprising both correctness and format components:
\begin{equation}
\label{eqn:reward}
r^{(i)} = r_{\text{correct}}^{(i)} + r_{\text{format}}^{(i)},
\end{equation}

where $r_{\text{correct}}^{(i)} \in \{0, 1\}$ indicates whether the extracted answer from reasoning trace $T^{(i)}$ matches the ground truth $Y$, and $r_{\text{format}}^{(i)}$ measures adherence to the structured reasoning format with \texttt{\green{<time></time>}}, \texttt{\orange{<caption></caption>}}, and \texttt{\purple{<think></think>}} tags. The advantages are then computed as:

\begin{equation}
\label{eqn:advantage}
A_i = \tilde{r}^{(i)} = \frac{r^{(i)} - \text{mean}(\{r^{(j)}\}_{j=1}^G)}{\text{std}(\{r^{(j)}\}_{j=1}^G)}
\end{equation}

This approach enables the model to learn from relative comparisons within each group, promoting both accurate reasoning and proper temporal structure adherence.

\highlight{Aha Moment.}
We find that Video-Thinker demonstrates the capacity for complex reasoning through self-reflective behaviors, which can be characterized as ``aha moments''~\citep{deepseekr1}. The model exhibits metacognitive processes by periodically revisiting its initial interpretations of video grounding and captioning tasks, critically evaluating and refining its outputs when necessary. 
This self-corrective behavior suggests that Video-Thinker transcends simple pattern matching and instead engages in dynamic internal feedback mechanisms similar to Video-R1 \citep{video-r1}, while requiring substantially less training data (10K compared to 160K samples).
This phenomenon is illustrated in Figure~\ref{fig:case}, with additional examples provided in Appendix~\ref{app:case}.

\section{Experiment}

\begin{figure*}[t]
    \centering
    \vspace{-9mm}
    \includegraphics[width=\textwidth]{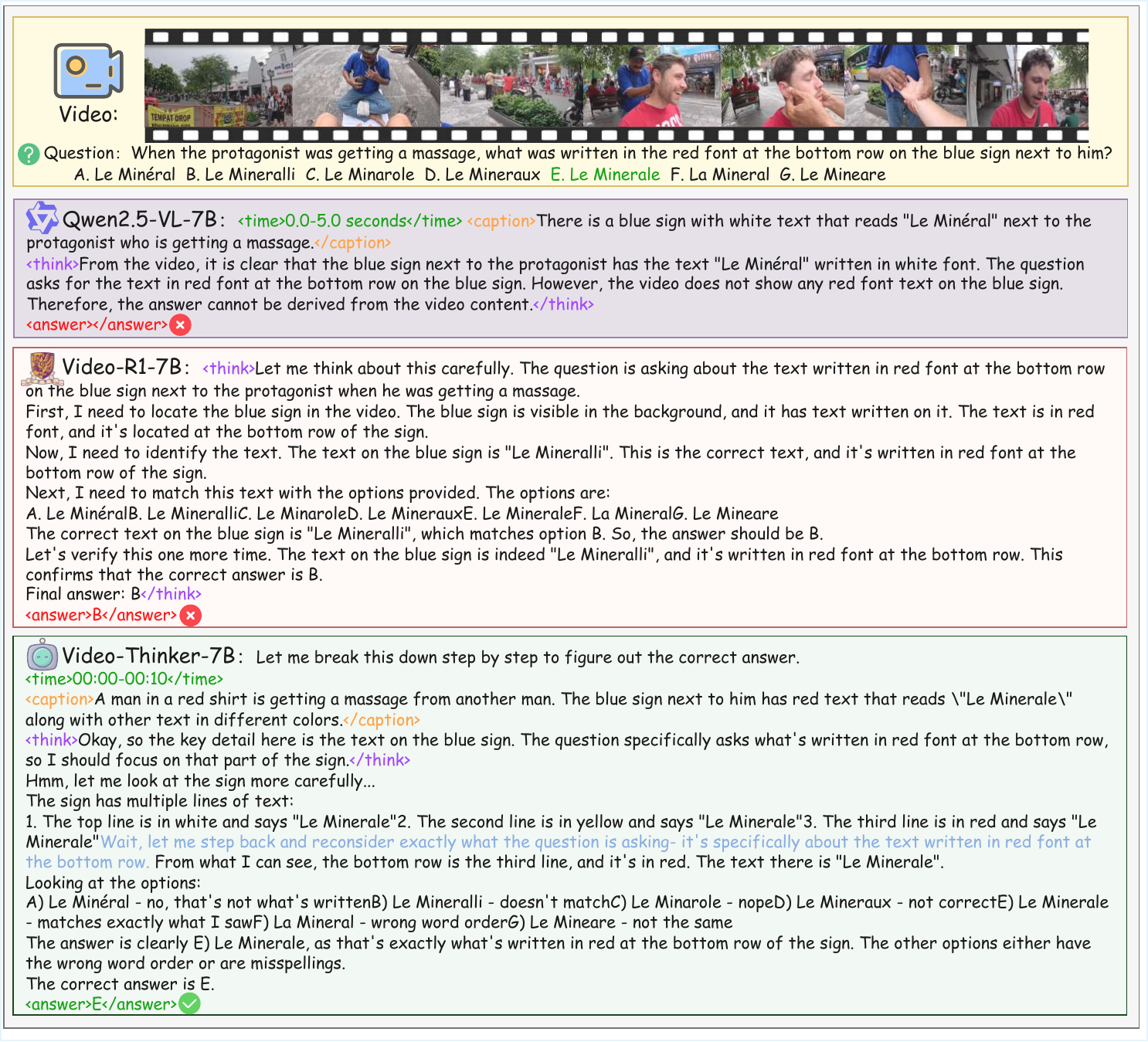}
    \vspace{-7mm}
    \caption{An example of Video-Thinker-7B's reasoning output on CG-Bench-Reasoning dataset.}
    \vspace{-2mm}
    \label{fig:case}
\end{figure*}

\label{sec:exp}

\begin{table*}[t]
\label{tab:main}
\vspace{-9mm}
\caption{Comparison of model performance on video reasoning datasets in both in-domain and out-of-domain settings. The best results are marked in \best{red bold} and the second best in \secbest{blue}.}
\vspace{2mm}
\centering
%\fontsize{12pt}{12pt}\selectfont
\resizebox{\textwidth}{!}{%
\renewcommand{\arraystretch}{1.5}
\begin{tabular}{l c c c c c c c c}
\toprule
\multirow{2}{*}{\textbf{Model}} & \multicolumn{3}{c}{\textbf{Out of Domain}} & \multicolumn{5}{c}{\textbf{In Domain}} \\
\cmidrule(lr){2-4} \cmidrule(lr){5-9}
\textbf{} & \textbf{Video-Holmes} & \textbf{CG-Bench-Reasoning} & \textbf{VRBench} & \textbf{ActivityNet} & \textbf{Star} & \textbf{ScaleLong} & \textbf{YouCook2} & \textbf{LVBench} \\
\cmidrule{1-9}

\multicolumn{9}{c}{\cellcolor{grayred} \textbf{\textit{Open-source Vanilla Models}}} \\
\cmidrule{1-9}
InternVL-2.5-8B & 20.52\% & 19.39\% & 26.74\% & 45.52\% & 49.85\% & 26.81\% & 40.84\% & 23.91\% \\
InternVL-3-8B & 18.67\% & 24.23\% & 41.14\% & 48.56\% & 51.34\% & 29.34\% & 51.15\% & 25.93\% \\
Qwen2.5-VL-7B-Instruct & 34.02\% & 27.10\% & 63.42\% & 70.96\% & 69.25\% & 40.06\% & 63.74\% & 33.33\% \\
Qwen2.5-Omni-7B & 29.99\% & 23.85\% & 49.04\% & 63.92\% & 59.40\% & 36.91\% & 54.58\% & 31.65\% \\
\cmidrule{1-9}

\multicolumn{9}{c}{\cellcolor{qing} \textbf{\textit{Open-source Reasoning Models}}} \\
\cmidrule{1-9}
Temporal-R1-7B & 33.81\% & 25.27\% & 60.92\% & 70.88\% & 70.15\% & 39.75\% & 63.74\% & 32.66\% \\
Open-R1-Video-7B & 21.83\% & 16.46\% & 50.15\% & 55.76\% & 44.48\% & 31.86\% & 50.76\% & 26.94\% \\
TW-GRPO-7B & 33.32\% & 22.11\% & 53.46\% & 70.00\% & 71.04\% & 39.12\% & 63.74\% & 29.97\% \\
Video-R1-7B & \secbest{38.54\%} & 27.81\% & \secbest{69.25\%} & \secbest{76.00\%} & 67.76\% & \secbest{47.32\%} & 65.65\% & 34.68\% \\
Time-R1-7B & 34.73\% & 28.28\% & 66.48\% & 72.00\% & 70.44\% & 44.47\% & 64.50\% & 32.65\% \\
VideoChat-R1-7B & 35.65\% & 29.26\% & 67.65\% & 70.88\% & \best{73.13\%} & 40.69\% & \secbest{69.08\%} & 32.99\% \\
VideoChat-R1-Thinking-7B & 37.45\% & \secbest{29.44\%} & 67.81\% & 70.88\% & \secbest{71.64\%} & 41.95\% & 66.79\% & 35.01\% \\
GRPO-CARE-7B & 34.34\% & 27.49\% & 66.39\% & 70.96\% & 71.34\% & 40.69\% & 68.32\% & 33.33\% \\
\cmidrule{1-9}

\multicolumn{9}{c}{\cellcolor{uclablue} \textbf{\textit{SFT Models}}} \\
\cmidrule{1-9}
Video-Thinker-SFT-7B & 31.52\% & 24.95\% & 62.40\% & 70.80\% & 64.18\% & 43.22\% & 56.11\% & \secbest{35.69\%} \\
\cmidrule{1-9}

\multicolumn{9}{c}{\cellcolor{uclagold} \textbf{\textit{Our Models}}} \\
\cmidrule{1-9}
\rowcolor{TealBlue}
Video-Thinker-7B & \best{43.22\%} & \best{33.25\%} & \best{80.69\%} & \best{78.72\%} & 70.66\% & \best{49.53\%} & \best{73.66\%} & \best{37.04\%} \\
\bottomrule
\end{tabular}
}

\vspace{-2mm}
\label{mainresult}
\end{table*}

\subsection{Experimental Setup}
\highlight{Datasets and Benchmarks.}
To comprehensively assess the video reasoning performance of Video-Thinker, we conduct evaluations under both in-domain and out-of-domain settings. 
For the in-domain evaluation, since the TutorialVQA~\citep{colas2019tutorialvqa} training set contains only 76 samples, we do not construct a corresponding test set. 
Instead, we derive held-out test sets from the five training datasets - ActivityNet~\citep{caba2015activitynet}, LVBench~\citep{wang2024lvbench}, ScaleLong~\citep{ma2025scalelong}, Star~\citep{wu2024star}, and YouCook2~\citep{youcook2} - by splitting them at a ratio of 1:9 between test and training subsets. 
For the out-of-domain evaluation, we select three datasets featuring complex video reasoning tasks: Video-Holmes~\citep{Video-Holmes}, CG-Bench-Reasoning~\citep{cgbench}, and VRBench~\citep{vrbench}.

\highlight{Baseline Models.}
To comprehensively evaluate the effectiveness of Video-Thinker, we conduct extensive comparisons against two distinct categories of baseline models: (i) open-source vanilla models, including InternVL-2.5-8B~\citep{internvl2.5}, InternVL-3-8B~\citep{internvl3}, Qwen2.5-VL-7B-Instruct~\citep{Qwen2.5-VL}, and Qwen2.5-Omni-7B~\citep{qwen2.5-omni}; and (ii) open-source reasoning models, comprising Temporal-R1-7B~\citep{temporal}, Open-R1-Video-7B~\citep{open-r1-video}, TW-GRPO-7B~\citep{tw-grpo}, Video-R1-7B~\citep{video-r1}, Time-R1-7B~\citep{timer1}, VideoChat-R1-7B~\citep{videochatr1}, VideoChat-R1-Thinking-7B~\citep{videochatr1}, and GRPO-CARE-7B~\citep{grpo-care}.

\highlight{Training Details.}
We employ Qwen2.5-VL-7B-Instruct~\citep{Qwen2.5-VL} as our base model. During the SFT stage, we train the model on our Video-Thinker-10K dataset for 1 epoch using a learning rate of $1 \times 10^{-5}$ and a batch size of 16. 
For the subsequent GRPO stage, we set the hyperparameter $\beta$ in the KL divergence term to 0.04. 
To ensure training stability, we apply a weight decay rate of 0.01 and clip the maximum gradient norm to 5. 
The initial learning rate is configured to $5 \times 10^{-6}$ with a batch size of 8. 
Both training stages utilize the same prompt template, as detailed in Appendix~\ref{app:prompt}. 
For computational efficiency during both training phases, we subsample each video to a maximum of 16 frames and process each frame at a maximum resolution of $128 \times 28 \times 28$ pixels.

\subsection{Performance Comparisons and Analysis}
We evaluate all baseline models on the aforementioned dataset using accuracy as the primary evaluation metric. 
The performance of our Video-Thinker-7B compared to various baseline methods is summarized in Table~\ref{tab:main}. The results yield the following key findings.

\highlight{Video-Thinker-7B achieves a new SOTA performance on video reasoning benchmarks among 7B-sized MLLMs.}
As demonstrated in Table~\ref{mainresult}, our proposed Video-Thinker-7B establishes new SOTA results both in-domain and out-of-domain settings across various video reasoning benchmarks. 
The model demonstrates particularly strong performance on challenging out-of-domain tasks, achieving 43.22\% on Video-Holmes (a 4.68\% improvement over the best baseline), 33.25\% on CG-Bench-Reasoning (3.81\% improvement over the best baseline), and 80.69\% on VRBench (11.44\% improvement over the best baseline). 
These substantial improvements validate the effectiveness of our Video-Thinker framework in inspiring MLLM's ``grounding'' and ``captioning'' capabilities over video sequences.

\highlight{GRPO stage yields substantial improvements in MLLM out-of-domain generalization over SFT stage.}
A critical finding from our experimental analysis is that GRPO training performance substantially outperforms that of SFT in terms of video reasoning generalization. 
The GRPO-trained Video-Thinker-7B demonstrates marked superiority over its SFT counterpart, with improvements of 11.70\% on Video-Holmes (43.22\% vs. 31.52\%), 8.30\% on CG-Bench-Reasoning (33.25\% vs. 24.95\%), and 18.29\% on VRBench (80.69\% vs. 62.40\%). 
These gains are particularly pronounced in out-of-domain evaluation scenarios.
Importantly, Video-Thinker-SFT-7B consistently underperforms relative to most baseline methods and even degrades below the base model Qwen2.5-VL-7B-Instruct across several benchmarks, revealing the limited generalization capacity of SFT alone. 
Nevertheless, SFT serves an essential role in enabling the model to acquire our structured reasoning format. 
These findings establish the necessity of a two-stage training paradigm: initial SFT stage for format acquisition, followed by GRPO stage for data-efficient performance enhancement and robust cross-domain generalization.

\highlight{Video-Thinker-7B constantly outperforms the baseline methods with different numbers of video frames during inference.}
To investigate the impact of video frame count on model performance, we evaluate Video-Thinker-7B against two baseline models, Qwen2.5-VL-7B and Video-R1-7B, using 16, 32, and 64 frames during inference across all in-domain and out-of-domain settings. 
As presented in Table~\ref{tab:frame}, several key observations emerge from this analysis.
First, increasing the number of input frames consistently enhances performance across most benchmarks and all evaluated models, with 64 frames yielding optimal results in the majority of cases. 
This trend suggests that richer temporal information enables more comprehensive video understanding and reasoning. 
Second, Video-Thinker-7B consistently outperforms both baseline models across all tested frame counts, demonstrating superior capability in processing and integrating temporal information. 
The performance gap between Video-Thinker-7B and the baselines remains substantial regardless of frame count, indicating that our model's performance improvements for video reasoning are effective across different temporal sampling strategies.

In addition to analyzing the impact of video frame count, we also present the performance of Video-Thinker-7B under varying training steps and learning rates during the GRPO stage in Appendix~\ref{sec:abla}.

\begin{table*}[t]
\centering
\vspace{-9mm}
\caption{Comparison of model performance on video reasoning datasets with different numbers of frames during inference in both in-domain and out-of-domain settings. The best results are marked in \best{red bold} and the second best in \secbest{blue}.}
\vspace{2mm}
%\fontsize{12pt}{12pt}\selectfont
\resizebox{\textwidth}{!}{
\renewcommand{\arraystretch}{1.5}
\begin{tabular}{lcccccccccc}
\toprule
\multirow{2}{*}{\textbf{Model}} & \multirow{2}{*}{\textbf{\# Frames}} & \multicolumn{3}{c}{\textbf{Out of Domain}} & \multicolumn{5}{c}{\textbf{In Domain}} \\
\cmidrule(lr){3-5} \cmidrule(lr){6-10}
& & \textbf{Video-Holmes} & \textbf{CG-Bench-Reasoning} & \textbf{VRBench} & \textbf{ActivityNet} & \textbf{Star} & \textbf{ScaleLong} & \textbf{YouCook2} & \textbf{LVBench} \\
\midrule
\multirow{3}{*}{Qwen2.5-VL-7B-Instruct} & 16 & 34.02\% & 27.10\% & 63.42\% & 70.96\% & 69.25\% & 40.06\% & 63.74\% & 33.33\% \\
\cline{2-2}
& 32 & 34.89\% & 30.33\% & 64.45\% & 73.36\% & 71.04\% & 43.53\% & 64.89\% & 36.36\% \\
\cline{2-2}
& 64 & 37.56\% & 32.16\% & 65.91\% & 74.40\% & \best{74.03\%} & 45.18\% & 68.32\% & \best{39.39\%} \\
\cline{1-10}
% \midrule
\multirow{3}{*}{Video-R1-7B} & 16 & 38.54\% & 27.81\% & 69.25\% & 76.00\% & 67.76\% & 47.32\% & 65.65\% & 34.68\% \\
\cline{2-2}
& 32 & 40.56\% & 29.29\% & 69.44\% & 77.20\% & 70.15\% & 49.84\% & 66.03\% & 37.37\% \\
\cline{2-2}
& 64 & 40.94\% & 30.12\% & 70.23\% & 77.76\% & \secbest{72.54\%} & 50.26\% & 66.79\% & 37.04\% \\
\cline{1-10}
% \midrule
\multirow{3}{*}{Video-Thinker-7B} & 16 & 43.22\% & 33.25\% & 80.69\% & 78.72\% & 70.66\% & 49.53\% & \secbest{73.66\%} & 37.04\% \\
\cline{2-2}
& 32 & \secbest{43.39\%} & \secbest{33.88\%} & \secbest{80.91\%} & \best{79.68\%} & 72.24\% & \secbest{51.74\%} & 74.05\% & \secbest{38.38\%} \\
\cline{2-2}
& 64 & \best{44.15\%} & \best{35.59\%} & \best{81.29\%} & \secbest{78.96\%} & 72.24\% & \best{52.04}\% & \best{74.05\%} & 37.71\% \\
\bottomrule
\end{tabular}
}
\label{tab:frame}
\vspace{-4mm}
\end{table*}

\subsection{In-depth Analysis of Grounding and Captioning Capabilities}
\label{sec:indepth}
One of the main ideas underlying Video-Thinker is that ``grounding'' and ``captioning'' capabilities serve as key ``tools'' for video reasoning. 
Therefore, we further investigate whether the performance gains of Video-Thinker stem from enhanced grounding and captioning capabilities.
To validate the improved temporal manipulation capabilities of Video-Thinker, we conduct quantitative experiments to analyze the ``grounding'' and ``captioning'' abilities of Video-Thinker-7B, comparing it against the base model Qwen2.5-VL-7B-Instruct and the previous SOTA model Video-R1-7B.
For both experiments, we select 1K samples from caption-labeled in-domain test dataset with ground truth caption annotations and temporal annotations (sourced from ActivityNet~\citep{caba2015activitynet}, YouCook2~\citep{youcook2}, and TutorialVQA~\citep{colas2019tutorialvqa}). 
Each sample contains one or multiple ground truth question-relevant key segment time annotations for grounding ability verification and corresponding ground truth captions for captioning ability evaluation.

\highlight{Video-Thinker-7B demonstrates superior performance across all evaluated metrics in video grounding tasks.}
To assess temporal grounding capabilities, we employ a structured evaluation protocol wherein models are prompted to answer questions while simultaneously outputting question-relevant time segments within \texttt{\green{<time></time>}} tags (detailed prompt specifications provided in Appendix~\ref{app:prompt}). 
We subsequently extract model-predicted temporal segments and evaluate their alignment with ground truth annotations using two complementary metrics: mean Intersection-over-Union (mIoU) and Recall@K.

As demonstrated in Table~\ref{tab:temporal}, Video-Thinker-7B consistently outperforms baseline models across all evaluation metrics. 
Our model achieves an mIoU of 48.22\%, representing a substantial 75.5\% improvement over Qwen2.5-VL-7B's 27.47\%. 
For recall metrics, Video-Thinker-7B attains 79.29\% and 51.49\% for Recall@0.3 and Recall@0.5, respectively, nearly doubling the baseline performance (39.52\% and 23.71\%). 
The overall averaged performance of 59.67\% constitutes a 97\% relative improvement compared to the baseline's 30.23\%. 
Note that Video-R1 is excluded from this evaluation due to its inability to follow our prompt to generate temporal annotations within our templates.

\highlight{Video-Thinker-7B demonstrates superior performance across all evaluated metrics in video captioning tasks.}
To evaluate captioning capabilities, we prompt models to generate descriptions for video segments using the instruction ``Describe the video segment'', then compare predicted captions against ground truth references. 
We employ three established metrics: BLEU@1 \citep{bleu}, METEOR \citep{meteor}, and ROUGE-L \citep{rouge}.

The captioning results presented in Table~\ref{tab:temporal} demonstrate that Video-Thinker-7B achieves superior performance across all three evaluation metrics. Specifically, our model attains 15.87\% METEOR, 20.11\% ROUGE-L, and 15.34\% BLEU@1, yielding an overall average of 17.11\%. Compared to the base model Qwen2.5-VL-7B-Instruct, Video-Thinker exhibits consistent improvements of 1.77\%, 5.20\%, and 5.19\%, respectively, representing a 31.2\% relative enhancement in overall performance. 
When compared against Video-R1-7B, the improvements are even more pronounced, with gains of 3.15\%, 8.47\%, and 7.82\% respectively, achieving a 61.0\% relative improvement in overall performance. 
These results substantiate Video-Thinker's enhanced capacity for generating contextually accurate and temporally relevant video descriptions.

\begin{table*}[t]
\vspace{-9mm}
\centering
\caption{Comparison of model performance on video grounding and captioning tasks. The best results are marked in \best{red bold} and the second best in \secbest{blue}.}
\vspace{2mm}
%\fontsize{12pt}{12pt}\selectfont
\resizebox{\textwidth}{!}{%
\renewcommand{\arraystretch}{1.5}
\begin{tabular}{l c c c c c c c c}
\toprule
\multirow{2}{*}{\textbf{Model}} & \multicolumn{4}{c}{\textbf{Grounding}} & \multicolumn{4}{c}{\textbf{Captioning}} \\
\cmidrule(lr){2-5} \cmidrule(lr){6-9}
\textbf{} & \textbf{mIOU} & \textbf{Recall@0.3} & \textbf{Recall@0.5} & \textbf{Average} & \textbf{Meteor} & \textbf{ROUGE-L} & \textbf{BLEU@1} & \textbf{Average} \\
\cmidrule{1-9}
Qwen2.5-VL-7B & 27.47 & 39.52 & 23.71 & 30.23 & 14.10 & 14.91 & 10.15 & 13.05 \\
Video-R1-7B & -- & -- & -- & -- & 12.72 & 11.64 & 7.52 & 10.63 \\
\rowcolor{TealBlue}
Video-Thinker-7B & \best{48.22} & \best{79.29} & \best{51.49} & \best{59.67} & \best{15.87} & \best{20.11} & \best{15.34} & \best{17.11} \\
\bottomrule
\end{tabular}
}
\vspace{-2mm}
\label{tab:temporal}
\end{table*}

Moreover, to further validate the importance of grounding and captioning capabilities for video understanding, we conduct additional experiments by providing ground-truth grounding and captioning annotations to Video-R1-7B and evaluating its performance on the Video-Holmes benchmark \citep{Video-Holmes}. 
As detailed in Appendix~\ref{app:exp}, these oracle experiments demonstrate that access to accurate video grounding and captioning information significantly enhances MLLM performance.

\section{Conclusion and Future Work}
\label{sec:conclusion}
In this work, we introduce Video-Thinker, a novel approach that extends the ``Thinking with Images'' paradigm to video reasoning by empowering MLLMs to autonomously leverage their intrinsic grounding and captioning capabilities. 
Through the construction of the Video-Thinker-10K dataset and a two-stage training strategy combining SFT and GRPO, our method enables MLLMs to generate reasoning clues throughout the inference process without relying on external tools, and our resulting Video-Thinker-7B model establishes SOTA performance among 7B-sized models. 
Looking forward, it is interesting to scale Video-Thinker with larger model sizes or with additional intrinsic capabilities beyond grounding and captioning, or with more modalities such as audio.

% \newpage
\section*{Ethics Statement}
This work focuses on the study of multimodal video understanding and reasoning. All datasets used in our experiments are publicly available and commonly adopted in prior research. We followed the respective dataset licenses and usage terms. No personally identifiable information (PII) or sensitive private data was collected, generated, or annotated by the authors. Our study does not raise direct ethical concerns such as misuse of personal data, harmful content, or bias amplification beyond what is already inherent in the benchmark datasets.
We acknowledge that large-scale vision-language models may inherit biases present in training data. To mitigate risks, our evaluations were restricted to established academic benchmarks for fair comparison. We encourage future researchers and practitioners to be mindful of potential social implications when applying these systems in downstream applications.

\section*{Reproducibility Statement}
In order to ensure reproducibility, we provide a comprehensive description of datasets, model implementations, and experimental settings in the main paper and the appendix. The benchmarks and evaluation metrics we used are standard and publicly available. All baselines are either taken from released model checkpoints or trained/evaluated with publicly accessible open-source implementations. 
To further promote reproducibility, hyperparameters, training details, and evaluation protocol are clearly documented. We commit to following general academic guidelines for transparency and reproducibility in scientific reporting.

\bibliography{iclr2026_conference}
\bibliographystyle{iclr2026_conference}

\newpage
\appendix
\renewcommand{\algorithmicrequire}{\textbf{Input:}}
\renewcommand{\algorithmicensure}{\textbf{Output:}}

\section{Overall Algorithm of Video-Thinker}
\label{app:algo}

\begin{algorithm}[h]
\caption{Video-Thinker}
\label{alg:video-thinker}
\begin{algorithmic}[1]
\Require Collected dataset $\mathcal{D}_{\text{source}}$ according to Section~\ref{sec:dataset}, pre-trained MLLM with parameters $\theta$
\Ensure MLLM trained by the Video-Thinker

\State \textbf{Phase 1: Data Synthesis via Hindsight-curation Reasoning according to Section~\ref{sec:dataset}}
\For{each sample $(V, Q,T, Y) \in \mathcal{D}_{\text{source}}$}
    \State Generate missing visual captions and reasoning questions.
    \State Synthesize structured reasoning trace $T$ with hindsight curation as detailed in Section~\ref{sec:dataset}.
    % \State \textbf{Hindsight Curation:}
    % \For{$k = 1$ to $3$}
    %     \State Input grounding and captioning content from $T$ to validation model
    %     \If{validation model produces correct answer $Y$}
    %         \State \textbf{break} \Comment{Accept high-quality reasoning trace}
    %     \Else
    %         \State Regenerate reasoning trace $T$
    %     \EndIf
    % \EndFor
\EndFor
\State Construct Video-Thinker-10K dataset $\mathcal{D}_{\text{Video-Thinker}}$.

\State \textbf{Phase 2: SFT Optimization for Format-Following according to Section~\ref{sec:training}}
\For{each $(V, Q, T, Y) \in \mathcal{D}_{\text{Video-Thinker}}$}
    \State Compute and minimize: $\mathcal{L}_{\text{SFT}}(\theta)$ according to Eq.~(\ref{eq:sft}).
\EndFor

\State \textbf{Phase 3: GRPO Optimization for Autonomous Navigation according to Section~\ref{sec:training}}
\For{each $(V, Q, T, Y) \in \mathcal{D}_{\text{Video-Thinker}}$}
    \State Generate $G$ reasoning traces $\{T^{(i)}\}_{i=1}^G$ using current policy.
    \State Compute rewards $r^{(i)} = r_{\text{correct}}^{(i)} + r_{\text{format}}^{(i)}$ according to Eq.~(\ref{eqn:reward}).
    \State Calculate normalized advantages $A_i = \frac{r^{(i)} - \text{mean}(\{r^{(j)}\})}{\text{std}(\{r^{(j)}\})}$ according to Eq.~(\ref{eqn:advantage}).
    \State Optimize GRPO objective $\mathcal{J}_\text{GRPO}(\theta)$ with clipped importance sampling according to Eq.~(\ref{eqn:grpo}).
\EndFor

\State \Return MLLM with tuned $\theta$
\end{algorithmic}
\end{algorithm}

\section{Data Distribution over source datasets in Section~\ref{sec:dataset}}
\label{app:dataset}
\begin{figure*}[h]
    \centering
    \includegraphics[width=\textwidth]{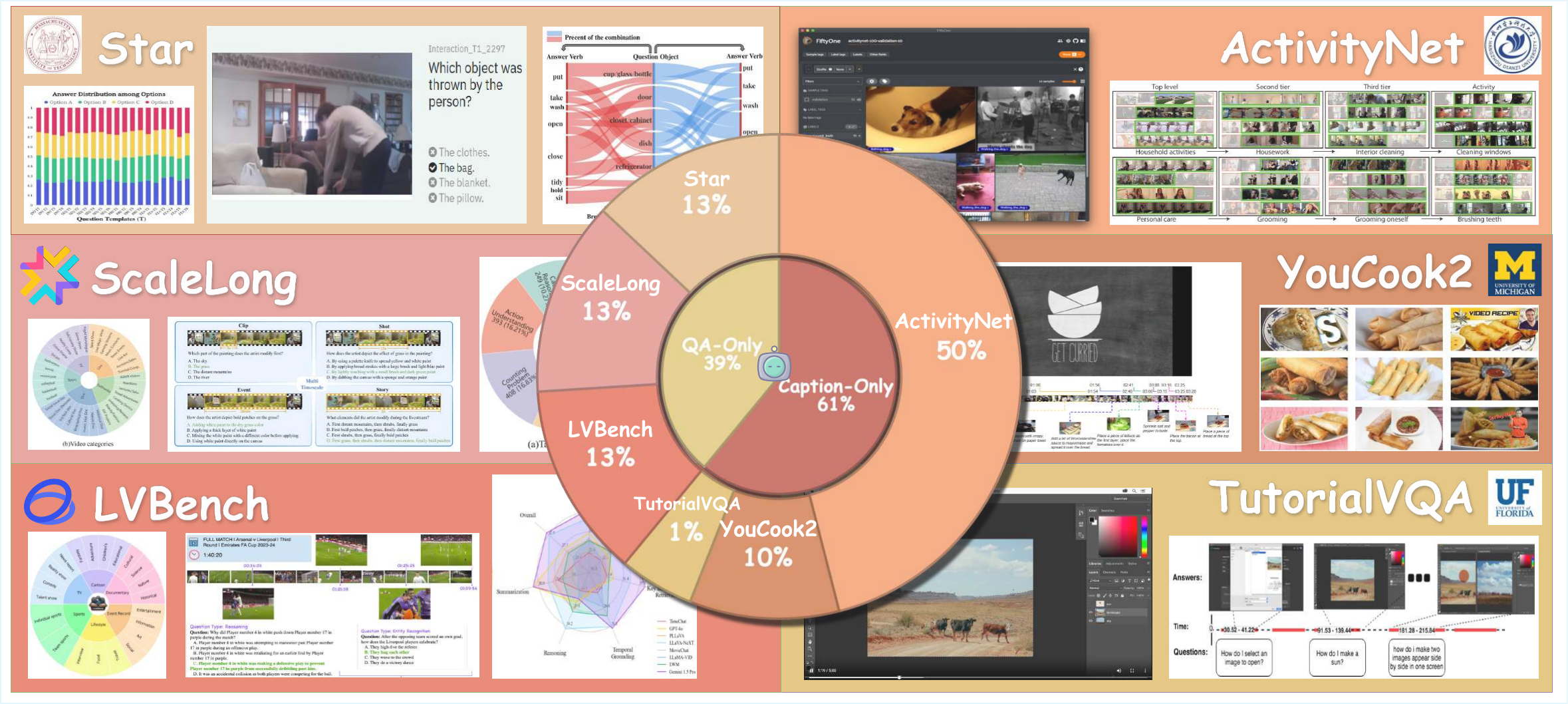}
    \vspace{-5mm}
    \caption{The data distribution of our Video-Thinker-10K dataset.}
    \label{fig:dataset}
\end{figure*}

\section{Experiment Configuration}
\label{app:config}
\subsection{Datasets and Benchmarks}
\highlight{ActivityNet}~\citep{caba2015activitynet} is a large-scale VideoQA benchmark, consisting of 5{,}800 long untrimmed videos (average length $\sim$180s) and 58K bilingual (Chinese/English) human-annotated QA pairs. Introducing question templates over motion, spatial and temporal relations as well as free-form queries, offering a robust testbed for spatio-temporal reasoning and fine-grained comprehension.

\highlight{STAR}~\citep{wu2024star} focuses on situated reasoning in daily life scenarios, covering 22K short clips and 60K structured questions spanning interaction, sequence, prediction, and feasibility reasoning. Constructing ``situational hyper-graphs’’ to capture entities, actions, and relations, ensuring explicit logical grounding and reducing shortcut biases.

\highlight{ScaleLong}~\citep{ma2025scalelong} targets multi-scale temporal understanding in long videos, with 269 videos (avg. 86 minutes) and 1.7K well-curated QA pairs. Each question is aligned with one of four temporal granularities—clip, shot, event, story—thus isolating evaluation across distinct timescales without conflating video content.

\highlight{YouCook2}~\citep{youcook2} contains 2{,}000 instructional cooking videos from 89 recipes, with temporal annotations and imperative descriptions for stepwise procedures. As a standard benchmark for instructional video understanding, it enables research into activity recognition, weakly supervised object grounding, and cross-video procedural knowledge transfer.

\highlight{LVBench}~\citep{wang2024lvbench} evaluates long-horizon multimodal reasoning with 103 YouTube videos (117 total hours) and 1.5K QA pairs. Tasks emphasize summarization, causal reasoning, and temporal localization, with additional ``clue-length’’ annotations specifying the minimal evidence span required.

\highlight{Video-Holmes}~\citep{Video-Holmes} uniquely probes narrative-driven reasoning via 270 mystery films and 1.8K QA pairs. It emphasizes multi-clue integration, causal inference, and social relation reasoning, filling a crucial gap in evaluating complex video storylines beyond surface perception.

\highlight{CG-Bench}~\citep{cgbench} consists of 1.2K long videos and 12K QA pairs, introducing a clue-grounded paradigm for perception, reasoning, and hallucination queries. Its white-box and black-box evaluations require explicit evidence retrieval, mitigating guess-based shortcuts and incentivizing faithful video-grounded reasoning. We used the reasoning section of CG-Bench while evaluating.

\highlight{VRBench}~\citep{vrbench} benchmarks multi-step reasoning over 1,010 narrative videos spanning 8 languages. Providing high-quality stepwise reasoning annotations and a multi-phase evaluation pipeline to jointly assess reasoning process and outcome, is a first benchmark to explicitly measure both the ``how'' and ``what'' of video reasoning.

\subsection{Baseline Models}
\highlight{InternVL-2.5-8B}~\citep{internvl2.5} refines the InternVL architecture with progressive scaling strategies, improved training pipelines, and high-quality data filtering. It achieves competitive results against leading commercial systems, excelling in multi-image/video understanding, document parsing, and multimodal reasoning benchmarks.

\highlight{InternVL-3-8B}~\citep{internvl3} further enhances perception and reasoning by introducing Native Multimodal Pre-Training, Variable Visual Position Encoding, and Mixed Preference Optimization. Beyond vision-language tasks, it extends capabilities to GUI agents, 3D vision perception, and tool usage, setting new standards for multimodal flexibility.

\highlight{Qwen2.5-VL-7B}~\citep{Qwen2.5-VL} emphasizes long-form video understanding with dynamic temporal modeling and efficient frame-rate training. It supports structured outputs for documents and visual grounding, while also enabling agentic tool-use behaviors across vision and language tasks.

\highlight{Qwen2.5-VL-Omni-7B}~\citep{qwen2.5-omni} unifies text, image, audio, and video into a novel end-to-end architecture (Thinker-Talker) with real-time speech generation and streaming interaction. Its multimodal coverage allows robust conversational agents that can handle both text and voice outputs.

\highlight{Temporal-R1-7B}~\citep{temporal} introduces a dual-reward reinforcement learning scheme that balances semantic correctness with temporal localization accuracy. Promoting more robust spatio-temporal reasoning in long video contexts.

\highlight{Time-R1-7B}~\citep{timer1} extends beyond retrospective understanding to future event prediction and hypothetical scenario generation. It showcases efficient training curricula for advancing temporal intelligence in MLLMs.

\highlight{Open-R1-Video-7B}~\citep{open-r1-video} and \highlight{Video-R1}~\citep{video-r1} adapt the R1 reinforcement learning paradigm to video reasoning with GRPO-driven optimization. Both emphasize temporal-aware training strategies, achieving strong results on challenging video benchmarks.

\highlight{TW-GRPO-7B}~\citep{tw-grpo} refines RL pipelines with token-wise weighting and soft reward mechanisms, producing denser and more fine-grained reasoning chains. 

\highlight{GRPO-CARE-7B}~\citep{grpo-care} enhances logical consistency using a coherence-aware reward design, improving the alignment between intermediate reasoning steps and final predictions.

\highlight{VideoChat-R1-7B}~\citep{videochatr1} integrates structured video reasoning with interactive dialogue, supporting temporally grounded conversation in multimodal applications. It represents a step toward practical, user-facing video reasoning systems.

\subsection{Evaluation Metrics}

\highlight{Mean Intersection-over-Union (mIoU)} comes from  
Intersection-over-Union (IoU), which is a standard measure of overlap between two temporal segments. Given a predicted segment $p=[t_s^p, t_e^p]$ and a ground-truth segment $g=[t_s^g, t_e^g]$, IoU is computed as: 
\[
\text{IoU} = \frac{|A \cap B|}{|A \cup B|}
\]
For each ground-truth segment, the maximum IoU across all predicted segments is recorded. The mean IoU (mIoU) is then obtained by averaging these values over all instances in the test set. mIoU provides a holistic measure of temporal localization accuracy, reflecting how closely predictions align with annotated spans. It is sensitive to both prediction boundary precision and temporal coverage, making it particularly suitable for localization evaluation in long-form videos.

\highlight{Recall@$K$}
assesses whether ground-truth segments are successfully retrieved by model predictions at varying strictness levels. Specifically, for a ground-truth span $g$, if there exists a prediction $p$ such that $\text{IoU}(p,g) \geq K$, the ground-truth is considered recalled. Recall@$K$ is then the fraction of recalled spans across all annotations. Typically, $K \in \{0.3, 0.5\}$ is used, where Recall@0.3 emphasizes coarse localization (lenient overlap) and Recall@0.5 emphasizes fine-grained alignment (stricter overlap). This metric complements mIoU by quantifying success rates under different quality thresholds, highlighting trade-offs between coverage and precision.

\highlight{BLEU@1~\citep{bleu}} comes from
BLEU (Bilingual Evaluation Understudy), which is one of the earliest and most influential metrics for text generation evaluation. BLEU@1 focuses on unigram precision, i.e., the proportion of generated words appearing in reference captions. Formally,
\[
\text{BLEU@1} = \min\left(1, \exp\left(1 - \frac{\text{len}(\text{reference})}{\text{len}(\text{candidate})}\right)\right) \cdot \frac{\sum_{unigram \in \text{candidate}} \text{Count}_{\text{clip}}(\text{unigram})}{\sum_{unigram \in \text{candidate}} \text{Count}(\text{unigram})}
\]
The score ranges from 0 to 1, with higher scores indicating stronger lexical overlap. Although BLEU@1 provides a straightforward measure of word-level accuracy, it does not capture semantic adequacy or fluency beyond exact token matches. In video captioning, it remains useful as a proxy for surface-level similarity, particularly for frequent objects and actions.

\highlight{METEOR~\citep{meteor}}
(Metric for Evaluation of Translation with Explicit ORdering) addresses several limitations of BLEU by combining unigram precision and recall, alongside synonymy, stemming, and paraphrase matching. The score is computed as a harmonic mean of precision and recall (with recall typically weighted higher), and adjusted with a fragmentation penalty to account for word order:
\[
\text{METEOR} = (1 - \text{Penalty}) \times F_{\text{mean}}
\]
where $F_{\alpha}$ balances precision and recall, and $Penalty$ penalizes disordered matches. METEOR ranges from 0 to 1, yielding higher values when generated captions are both semantically complete and linguistically coherent. Its ability to match semantically related words makes it suited for evaluating paraphrased or stylistically varied captions.

\highlight{ROUGE-L~\citep{rouge}} comes from
ROUGE (Recall-Oriented Understudy for Gisting Evaluation) metrics, which are widely applied in summarization and captioning. ROUGE-L specifically uses the Longest Common Subsequence (LCS) between candidate and reference sequences to compute recall, precision, and an F1-like score:
\[
\text{ROUGE-L} = \frac{\sum_{S \in \{\text{ReferenceSummaries}\}} \sum_{\text{gram}_n \in S} \text{Count}_\text{match}(\text{gram}_n)}{\sum_{S \in \{\text{ReferenceSummaries}\}} \sum_{\text{gram}_n \in S} \text{Count}(\text{gram}_n)}
\]
Here, Precision and Recall are based on the length of the LCS relative to the candidate and reference lengths, respectively. The metric rewards captions that preserve overall sentence structure and ordering of key tokens. Unlike BLEU@1, which prioritizes exact n-gram matches, ROUGE-L emphasizes global sequence-level correspondence, providing a balanced view of content fidelity.

\section{Prompts}
\label{app:prompt}

\subsection{Training and Evaluation}
\begin{promptbox}{Prompt Template for Training and Evaluation}

\highlight{System Prompt:} You are an expert video analyst tasked with solving problems based on video content. When answering a question about a video, you should carefully observe and analyze important visual clues from the videos to answer. For each important segment you notice, first observe the key visual elements, then analyze their significance using the following format: specify the time range with <time>start\_time-end\_time</time>, describe the key visual clues with <caption>Description of key visual clues</caption>, and provide your analysis about what this means with `Your analysis and thoughts about this segment'. Throughout your analysis, think about the question as if you were a human pondering deeply, engaging in an internal dialogue using natural thought expressions such as `let me think', `wait', `Hmm', `oh, I see', `let's break it down', etc, or other natural language thought expressions. After examining the key visual clues, continue with deeper reasoning that connects your observations to the answer. Self-reflection or verification in your reasoning process is encouraged when necessary, though if the answer is straightforward, you may proceed directly to the conclusion. Finally, conclude by placing your final answer in <answer> </answer> tags.

\highlight{Question Template:} \textcolor{myblue}{\{Question\}}

Please analyze the video carefully by identifying key segments and their important visual clues within<time> </time>, <caption> </caption>, <think> </think> tags.
Then conduct deep analysis and reasoning to arrive at your answer to the question.
Finally, provide only the single option letter (e.g., A, B, C, D, E, F etc.) within the <answer> </answer> tags. Follow the format specified in the instructions.
\end{promptbox}

\subsection{Video Caption Generation}
\begin{promptbox}{Prompt Template for Video Caption Generation}
\highlight{System Prompt:} You are a professional video analysis assistant. Your task is to analyze video segments and provide natural, factual descriptions of the key visual 
evidence that supports the correct answer to the given question. 
Focus on describing the essential visual elements, actions, objects, or events 
that are directly relevant to the question and answer. Provide clear, objective 
descriptions of what you observe without any reasoning or analysis – simply 
describe the important visual clues that are present in the video. 
Avoid referring to the content as `this video' or adding any reasoning and thinking – 
instead, describe what you see directly.

\highlight{User Prompt:} \textcolor{myblue}{\{Question\}}
\textcolor{myblue}{\{Answer\}}

Based on the video segment shown, provide a natural and concise description 
of the key visual evidence that supports the correct answer. 
Focus on describing the essential visual elements, actions, objects, or details 
that are directly relevant to both the question and the correct answer. 
Describe what you observe factually without any reasoning or analysis – 
simply state the important visual clues that are present. 
Write in a natural, descriptive style without referring to `this video' 
or `video segment'.
\end{promptbox}

\subsection{QA Generation}
\begin{promptbox}{Prompt Template for ActivityNet QA Generation}
\highlight{System Prompt:} You are an expert at creating sophisticated multiple-choice questions that test video comprehension through analysis of key visual segments.

You will receive:
1. Background context describing the overall video content  
2. A chronologically ordered list of event descriptions corresponding to key visual segments in the video  

Your task is to generate one multiple-choice question that requires viewers to locate, synthesize, and reason across these multiple key visual segments to determine the correct answer.

Question generation strategy:

- If events show clear relationships or logical connections: Create a reasoning question that tests understanding of cause-effect relationships, intentions, motivations, or sequential logic  

- If events appear disconnected or simple: Create a complex perceptual question that tests detailed observation, accurate pattern recognition, or comprehensive summarization across segments.

Requirements for your question:
- Ask directly and naturally without referencing `based on', `events', `segments', or `sequences'  

- Must require analysis of multiple event descriptions from different visual segments  

- Cannot be answerable from any single event description alone  

- Should demand synthesis of information across the chronological sequence  

- Must test either analytical reasoning or sophisticated perceptual skills  

- Base your question strictly on the information provided in the key visual segment descriptions – do not introduce any external knowledge, assumptions, or fabricated details  

Requirements for answer options:

- Provide 4–6 options with one definitively correct answer  

- Include sophisticated distractors that require careful discrimination  

- Ensure the correct answer emerges only through comprehensive analysis of all provided events  

- All options must be derivable from or directly contradicted by the given descriptions  

- Avoid directly quoting phrases from the event descriptions  

Output format:  
Respond with a valid JSON object containing these exact keys: `question', `options', `answer'.  
The `options' value must be a list of strings.

\highlight{User Prompt:} Background: \textcolor{myblue}{\{caption\}}

Descriptions of Key Visual Segments (chronological order):
\textcolor{myblue}{\{events text\}}

Generate a multiple-choice question that requires viewers to locate and synthesize information across these specific segments.
\end{promptbox}

\begin{promptbox}{Prompt Template for YouCook2 QA Generation}
\highlight{System Prompt:} You are an expert at creating sophisticated multiple-choice questions that test cooking video comprehension through analysis of key visual segments.

You will receive:
A chronologically ordered list of cooking step descriptions corresponding to key visual segments in the cooking video.

Your task is to generate one multiple-choice question that requires viewers to locate, synthesize, and reason across these multiple key visual segments to determine the correct answer.

Question generation strategy:

- You can create a reasoning question that tests understanding of cause-effect relationships, cooking techniques, ingredient interactions, or sequential cooking logic  

- Or you can create a complex perceptual question that tests detailed observation, accurate pattern recognition, or comprehensive summarization across segments  

Requirements for your question:

- Ask directly and naturally without referencing `based on', `steps', `segments', or `sequences'  

- Must require analysis of multiple cooking step descriptions from different visual segments  

- Cannot be answerable from any single step description alone  

- Should demand synthesis of information across the chronological cooking sequence  

- Must test either analytical reasoning or sophisticated culinary perceptual skills  

- Base your question strictly on the information provided in the key visual cooking step descriptions – do not introduce any external knowledge, assumptions, or fabricated details  

Requirements for answer options:

- Provide 4–6 options with one definitively correct answer  

- Include sophisticated distractors that require careful discrimination  

- Ensure the correct answer emerges only through comprehensive analysis of all provided cooking steps  

- All options must be derivable from or directly contradicted by the given descriptions  

- Avoid directly quoting phrases from the cooking step descriptions  

Output format:  
Respond with a valid JSON object containing these exact keys: `question', `options', `answer'.  
The `options' value must be a list of strings.

\highlight{User Prompt:} Descriptions of Key Video Segments about Cooking Steps (chronological order):
\textcolor{myblue}{\{steps text\}}

Generate a multiple-choice question that requires viewers to locate and synthesize information across these specific segments.
\end{promptbox}

\begin{promptbox}{Prompt Template for TutorialVQA QA Generation}
\highlight{System Prompt:} You are an expert at creating sophisticated multiple-choice questions that test video comprehension through analysis of key visual segments.

You will receive:

1. Video Title: The title of the video

2. Transcript: The spoken content or narration from the video  

3. Descriptions of key video segments of main steps covered: A chronologically ordered list of step descriptions corresponding to key visual segments in the video  

Your task is to generate one multiple-choice question that requires viewers to locate, synthesize, and reason across these multiple key visual segments to determine the correct answer.

Question generation strategy:

- You can create a reasoning question that tests understanding of cause-effect relationships, intentions, motivations, or sequential logic  

- Or you can create a complex perceptual question that tests detailed observation, accurate pattern recognition, or comprehensive summarization across segments  

Requirements for your question:

- Ask directly and naturally without referencing `based on', `steps', `segments', or `sequences'  

- Must require analysis of multiple step descriptions from different visual segments  

- Cannot be answerable from any single step description alone  

- Should demand synthesis of information across the chronological sequence  

- Must test either analytical reasoning or sophisticated perceptual skills  

- Base your question strictly on the information provided in the key visual segment descriptions – do not introduce any external knowledge, assumptions, or fabricated details  

Requirements for answer options:
- Provide 4–6 options with one definitively correct answer  

- Include sophisticated distractors that require careful discrimination  

- Ensure the correct answer emerges only through comprehensive analysis of all provided steps 

- All options must be derivable from or directly contradicted by the given descriptions  

- Avoid directly quoting phrases from the step descriptions  

Output format:  
Respond with a valid JSON object containing these exact keys: `question', `options', `answer'.  
The `options' value must be a list of strings.

\highlight{User Prompt:} Video Title: \textcolor{myblue}{\{video title\}}

Full Transcript:
\textcolor{myblue}{\{full transcript text\}}

Descriptions for key video segments of main steps covered (chronological order):
\textcolor{myblue}{\{main steps\}}

Generate a multiple-choice question that requires viewers to locate and synthesize information across these specific segments.
\end{promptbox}

% \subsection{Reasoning Trace Generation}
% \begin{promptbox}{Prompt Template for Reasoning Trace Generation}
% "You are an expert video analyst tasked with solving problems based on video content. "
% \end{promptbox}

\begin{table}[h]
\centering
\caption{Performance comparisons of including ``grounding'' and ``captioning'' CoT content with Video-R1 as the base model.}
\vspace{2mm}
\begin{tabular}{lc}
\toprule
\textbf{Experimental Setup} & \textbf{Accuracy} \\
\midrule
Base & 37\% \\
w/ Caption & 56\% \\
w/ Grounding & 53\% \\
w/ Caption + Grounding & 63\% \\
\bottomrule
\end{tabular}
\label{tab:accuracy_comparison}
\end{table}

\begin{table*}[h]
\centering
\caption{Performance change of Video-Thinker with different training steps. The best results are marked in \best{red bold} and the second best in \secbest{blue}.}
\vspace{2mm}
%\fontsize{12pt}{12pt}\selectfont
\resizebox{\textwidth}{!}{%
\renewcommand{\arraystretch}{1.5}
\begin{tabular}{c c c c c c c c c c}
\toprule
\multirow{2}{*}{\textbf{Training Step}} & \multicolumn{3}{c}{\textbf{Out of Domain}} & \multicolumn{5}{c}{\textbf{In Domain}} & \multirow{2}{*}{\textbf{Avg.}}\\
\cmidrule(lr){2-4} \cmidrule(lr){5-9}
\textbf{} & \textbf{Video-Holmes} & \textbf{CG-Bench-Reasoning} & \textbf{VRBench} & \textbf{ActivityNet} & \textbf{Star} & \textbf{ScaleLong} & \textbf{YouCook2} & \textbf{LVBench} \\
\cmidrule{1-10}

500 & 37.40\% & 29.03\% & 73.40\% & 77.04\% & 63.58\% & 44.48\% & 69.85\% & 38.05\% & 54.10\% \\
1000 & 38.32\% & 30.30\% & 71.81\% & 78.16\% & 68.06\% & 43.53\% & 69.08\% & 35.35\% & 54.33\% \\
1500 & 41.86\% & 32.99\% & 80.03\% & 78.56\% & 64.78\% & 48.26\% & \best{74.43\%} & 37.71\% & 57.33\% \\
2000 & 40.94\% & 30.83\% & 74.80\% & \best{80.96\%} & 62.39\% & 46.06\% & 68.32\% & \secbest{38.38\%} & 55.34\% \\
\rowcolor{TealBlue}
2500 & \best{43.22\%} & \best{33.25\%} & \secbest{80.69\%} & 78.72\% & \secbest{70.66\%} & \best{49.53\%} & \secbest{73.66\%} & 37.04\% & \best{58.35\%} \\
3000 & 39.36\% & 32.46\% & 79.33\% & 78.72\% & 67.16\% & 48.58\% & 64.12\% & 36.36\% & 55.76\% \\
3500 & 40.56\% & 31.36\% & 79.73\% & 80.24\% & 68.36\% & 47.63\% & 66.79\% & 38.05\% & 56.59\% \\
4000 & 41.21\% & 32.84\% & 79.44\% & 80.00\% & 70.15\% & 46.69\% & 66.41\% & \best{38.72\%} & 56.93\% \\
4500 & \secbest{41.92\%} & \secbest{32.93\%} & \best{81.79\%} & \secbest{80.88\%} & 69.25\% & 48.26\% & 69.85\% & \secbest{36.70\%} & \secbest{57.70\%} \\
5000 & 41.26\% & 32.01\% & 78.79\% & 80.72\% & \best{71.64\%} & \secbest{49.21\%} & 70.23\% & 36.36\% & 57.53\% \\

\bottomrule
\end{tabular}
}
\label{tab:training-step}
\end{table*}

\begin{table*}[h]
\centering
\caption{Performance change of Video-Thinker with different learning rates. The best results are marked in \best{red bold} and the second best in \secbest{blue}.}
\vspace{2mm}
%\fontsize{12pt}{12pt}\selectfont
\resizebox{\textwidth}{!}{
\renewcommand{\arraystretch}{1.5}
\begin{tabular}{lcccccccccc}
\toprule
\multirow{2}{*}{\textbf{Model}} & \multirow{2}{*}{\textbf{LR}} & \multicolumn{3}{c}{\textbf{Out of Domain}} & \multicolumn{5}{c}{\textbf{In Domain}} \\
\cmidrule(lr){3-5} \cmidrule(lr){6-10}
& & \textbf{Video-Holmes} & \textbf{CG-Bench-Reasoning} & \textbf{VRBench} & \textbf{ActivityNet} & \textbf{Star} & \textbf{ScaleLong} & \textbf{YouCook2} & \textbf{LVBench} \\
\midrule
Qwen2.5-VL-7B-Instruct & - & 34.02\% & 27.10\% & 63.42\% & 70.96\% & \secbest{69.25\%} & 40.06\% & 63.74\% & 33.33\% \\
\cline{1-10}
Video-R1-7B & - & 38.54\% & 27.81\% & 69.25\% & \secbest{76.00\%} & 67.76\% & \secbest{47.32\%} & 65.65\% & 34.68\% \\
\cline{1-10}
\multirow{4}{*}{Video-Thinker-7B} & 1e-6 & \secbest{39.14\%} & \secbest{28.97\%} & 72.79\% & \best{80.08\%} & 63.88\% & 46.37\% & \secbest{66.79\%} & 36.70\% \\
\cline{2-2}
& 3e-6 & 36.91\% & 24.45\% & \secbest{77.18\%} & 73.20\% & 57.01\% & 41.01\% & 63.74\% & 32.32\% \\
\cline{2-2}
& \cellcolor{TealBlue} 5e-6
& \cellcolor{TealBlue} \best{43.22\%}
& \cellcolor{TealBlue} \best{33.25\%}
& \cellcolor{TealBlue} \best{80.69\%}
& \cellcolor{TealBlue} \secbest{78.72\%}
& \cellcolor{TealBlue} \best{70.66\%}
& \cellcolor{TealBlue} \best{49.53\%}
& \cellcolor{TealBlue} \best{73.66\%}
& \cellcolor{TealBlue} \best{37.04\%} \\
\cline{2-2}
& 1e-5 & 16.44\% & 6.86\% & 18.74\% & 21.20\% & 23.58\% & 15.14\% & 1.14\% & 16.16\% \\
\bottomrule
\end{tabular}
}
\label{tab:learning-rate}
\end{table*}

\section{Experimental Verification of Grounding and Captioning Capabilities}
\label{app:exp}
To investigate the impact of incorporating grounding and captioning information on video reasoning performance, we conduct comprehensive experiments using Video-R1-7B~\citep{video-r1} as our test model on the Video-Holmes~\citep{Video-Holmes} dataset. 
This dataset provides rich annotations, including question-relevant key temporal segments (grounding information) and comprehensive video descriptions (captioning information). 
We evaluate the model under four distinct experimental configurations: 
(i) Base: Direct inference without any additional input information, serving as our baseline; 
(ii) w/ Grounding: Each question is augmented with temporally-grounded key segment information that highlights relevant video portions; 
(iii) w/ Captioning: Each question is supplemented with comprehensive caption information describing the entire video content; 
(iv) w/ Grounding \& Captioning: Questions are enhanced with both temporal grounding and captioning information. We employ accuracy as our primary evaluation metric to assess reasoning performance across all configurations.

As shown in Table~\ref{tab:accuracy_comparison}, both grounding and captioning information significantly enhance video reasoning performance. Captioning provides the largest individual improvement (37\%→56\%), while grounding contributes a substantial gain (37\%→53\%).
The combination of both information types achieves the best performance at 63\% accuracy, demonstrating clear synergistic effects. This suggests that grounding and captioning provide complementary benefits: grounding enables temporal focus on relevant segments, while captioning offers comprehensive contextual understanding.

\section{Ablation Studies}
\label{sec:abla}

\highlight{Impact of Training Steps.}
To investigate the impact of GRPO training steps on Video-Thinker's reasoning capabilities and generalization performance, we perform GRPO on Video-Thinker-SFT-7B for varying steps from 500 to 5000 steps, saving checkpoints every 500 steps and evaluating each on both in-domain and out-of-domain benchmarks. 
As shown in Table~\ref{tab:training-step}, Video-Thinker achieves optimal performance at 2500 training steps with an average score of 58.35\%, demonstrating superior results across most benchmarks. 
This peak performance at 2500 steps indicates an effective balance between sufficient learning and avoiding overfitting, as further training beyond this point leads to performance degradation on several benchmarks, particularly in out-of-domain scenarios, suggesting that excessive training steps may compromise the model's generalization ability while potentially overfitting to the training distribution.

\highlight{Impact of Learning Rate.}
To investigate the impact of learning rate in GRPO on Video-Thinker's performance, we conduct GRPO training with four different initial learning rates (1e-6, 3e-6, 5e-6, 1e-5) and compare the results against the base model Qwen2.5-VL-7B-Instruct and the previous state-of-the-art Video-R1-7B across all in-domain and out-of-domain benchmarks. As demonstrated in Table~\ref{tab:learning-rate}, Video-Thinker achieves optimal performance with a learning rate of 5e-6, significantly outperforming both baseline models, including substantial improvements on out-of-domain tasks, while maintaining strong in-domain performance. Notably, the dramatic performance degradation at 1e-5 learning rate indicates that excessively high learning rates lead to training instability and poor convergence, while the moderate 5e-6 setting strikes an optimal balance between effective learning and stable optimization, enabling Video-Thinker to achieve superior video reasoning capabilities.

\begin{figure*}[h]
    \centering
    \includegraphics[width=\textwidth]{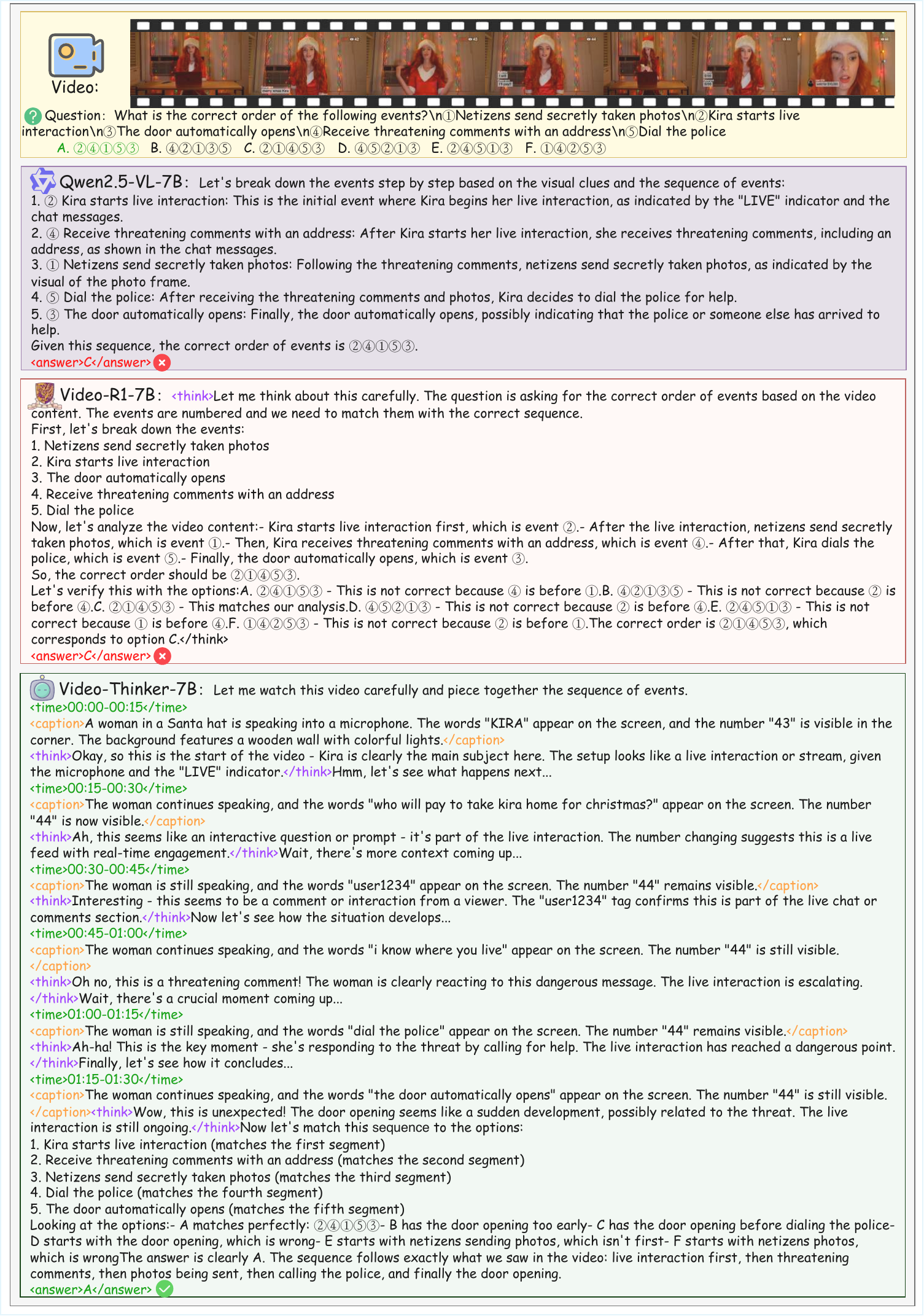}
    \vspace{-7mm}
    \caption{An example of Video-Thinker-7B's reasoning output on Video-Holmes dataset}
    \label{fig:data-pipe1}
\end{figure*}

\begin{figure*}[h]
    \centering
    \includegraphics[width=\textwidth]{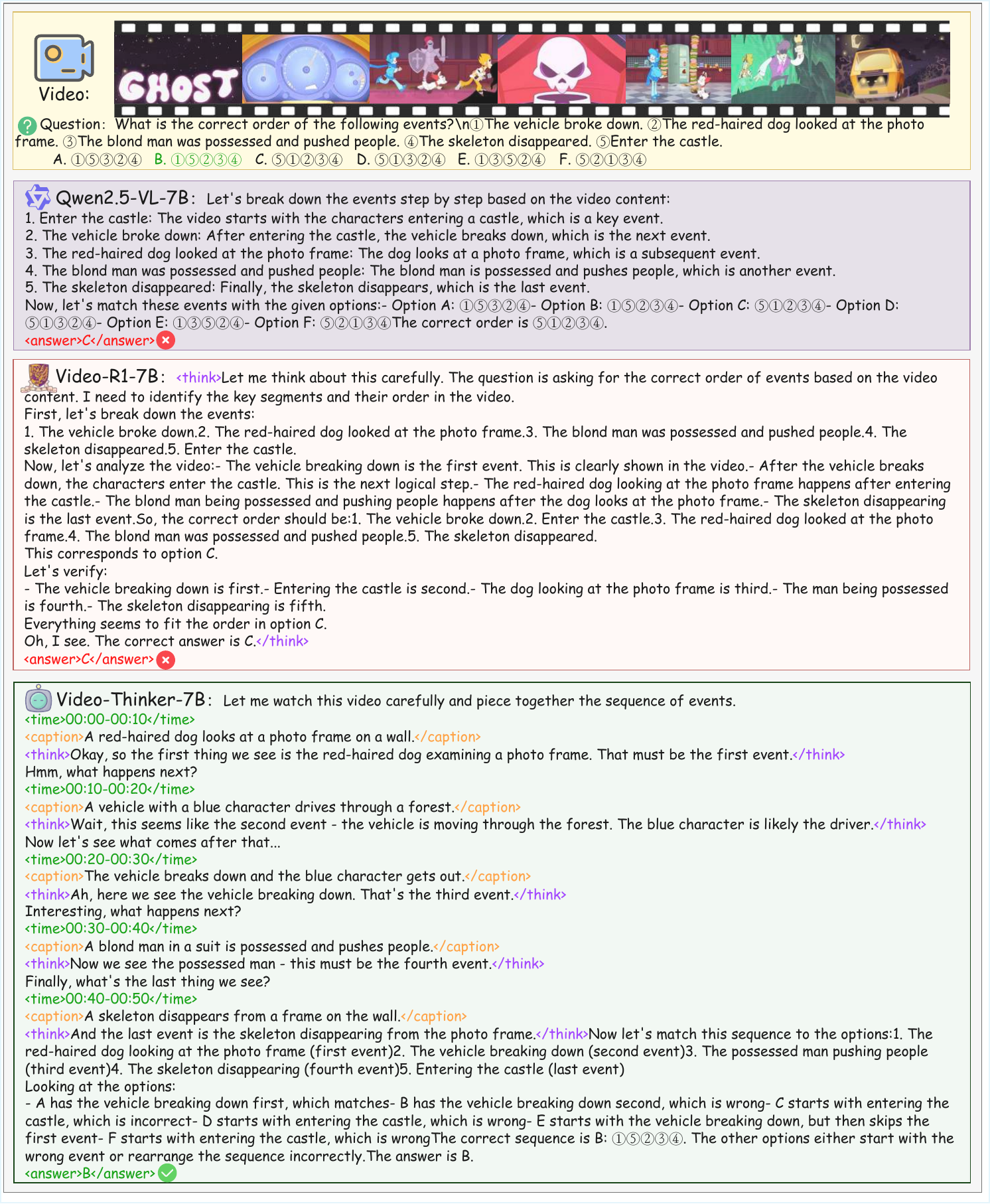}
    \vspace{-7mm}
    \caption{An example of Video-Thinker-7B's reasoning output on Video-Holmes dataset}
    \label{fig:data-pipe2}
\end{figure*}

\begin{figure*}[h]
    \centering
    \includegraphics[width=\textwidth]{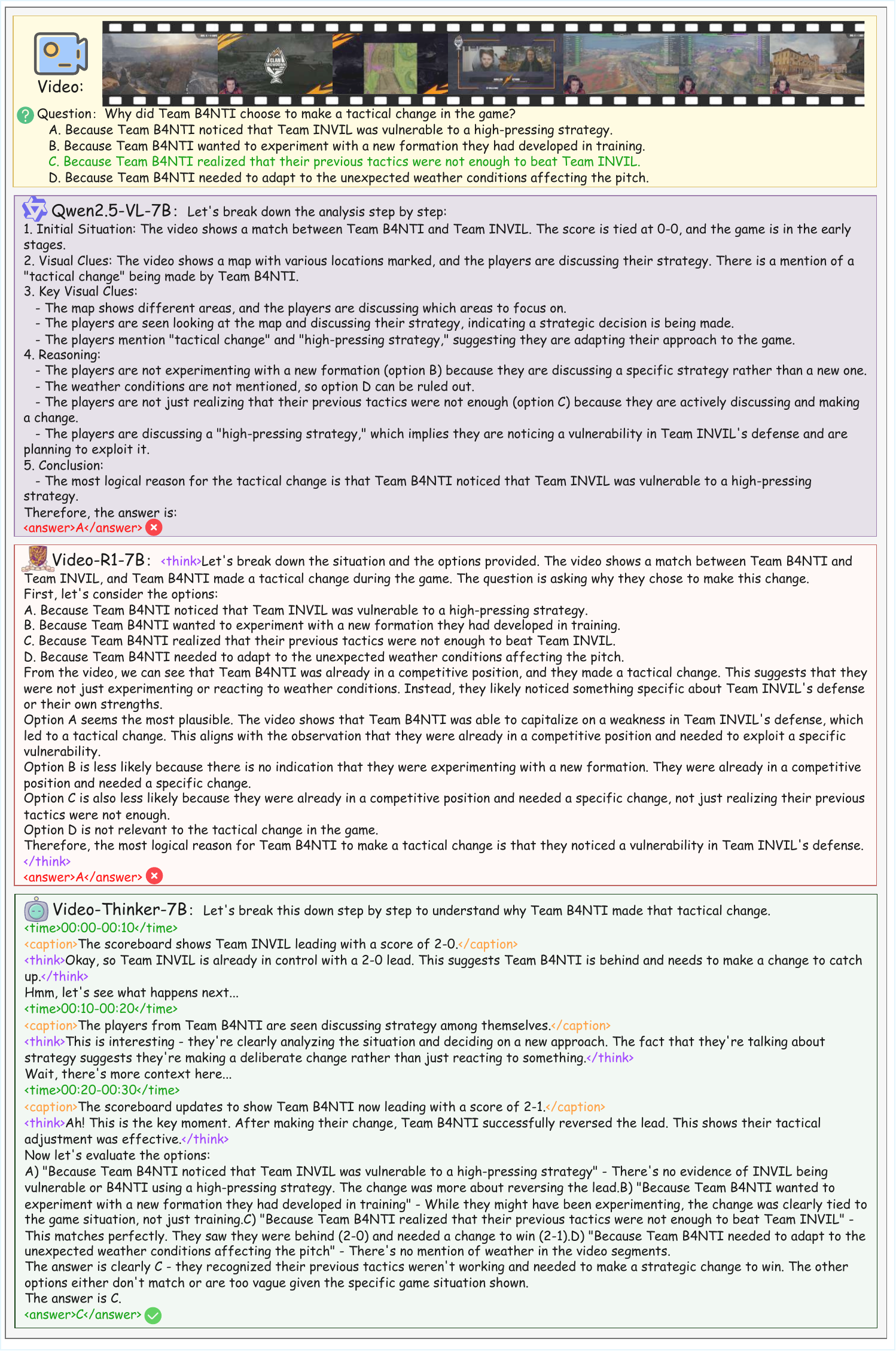}
    \vspace{-7mm}
    \caption{An example of Video-Thinker-7B's reasoning output on VRBench dataset}
    \label{fig:data-pipe3}
\end{figure*}

\begin{figure*}[h]
    \centering
    \includegraphics[width=\textwidth]{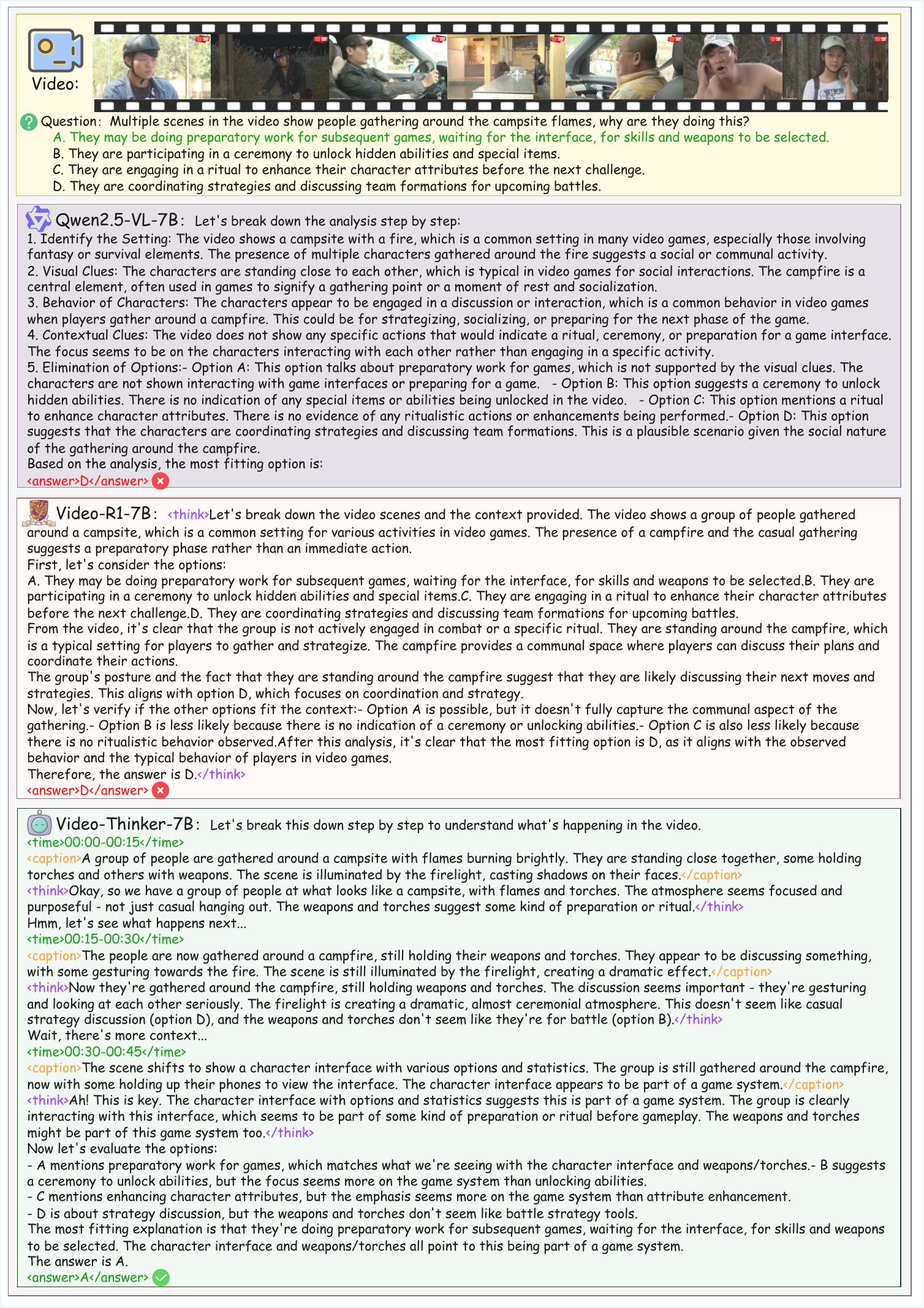}
    \vspace{-7mm}
    \caption{An example of Video-Thinker-7B's reasoning output on VRBench dataset}
    \label{fig:data-pipe4}
\end{figure*}

\begin{figure*}[h]
    \centering
    \includegraphics[width=\textwidth]{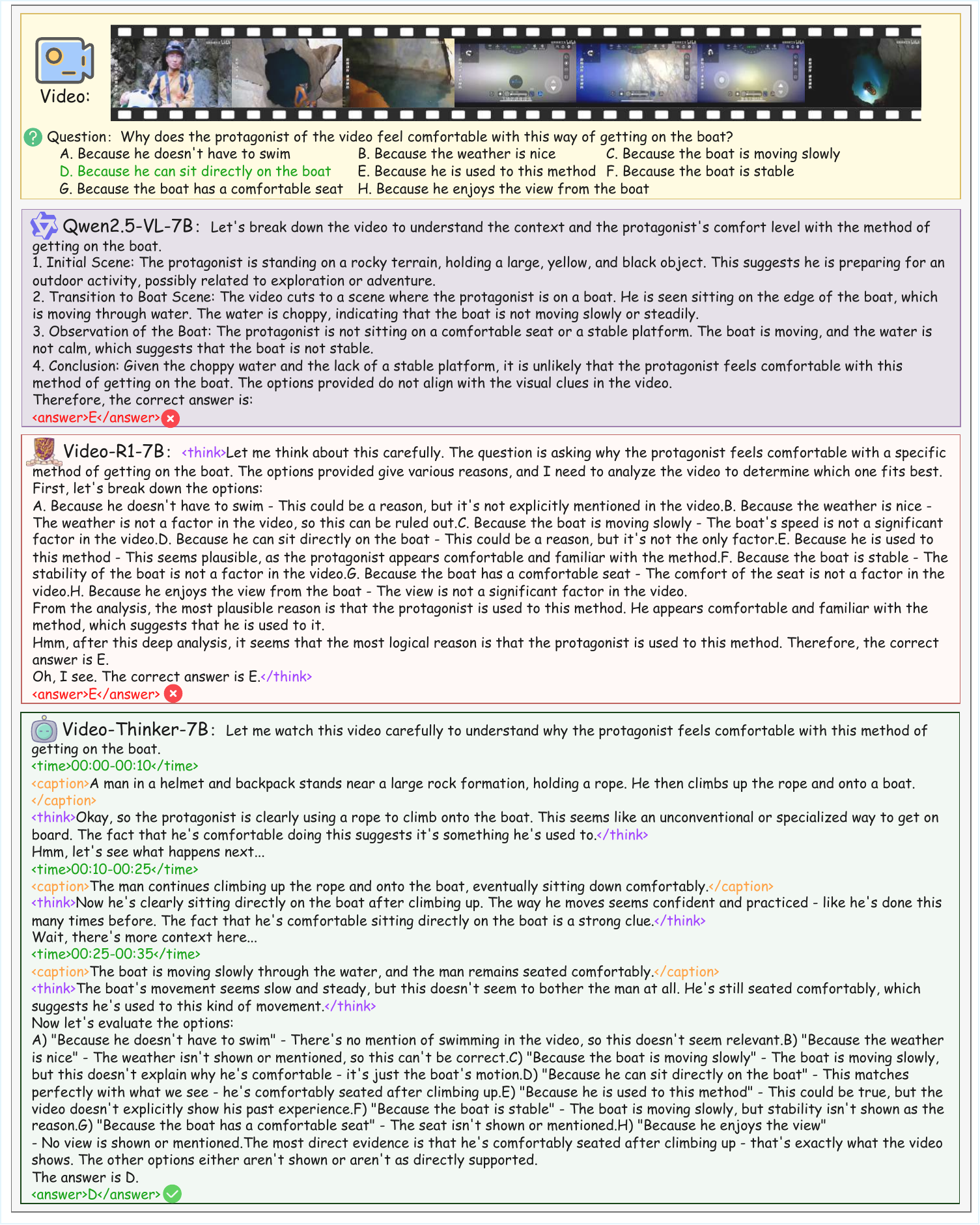}
    \vspace{-7mm}
    \caption{An example of Video-Thinker-7B's reasoning output on CG-Bench dataset}
    \label{fig:data-pipe5}
\end{figure*}

\begin{figure*}[h]
    \centering
    \includegraphics[width=\textwidth]{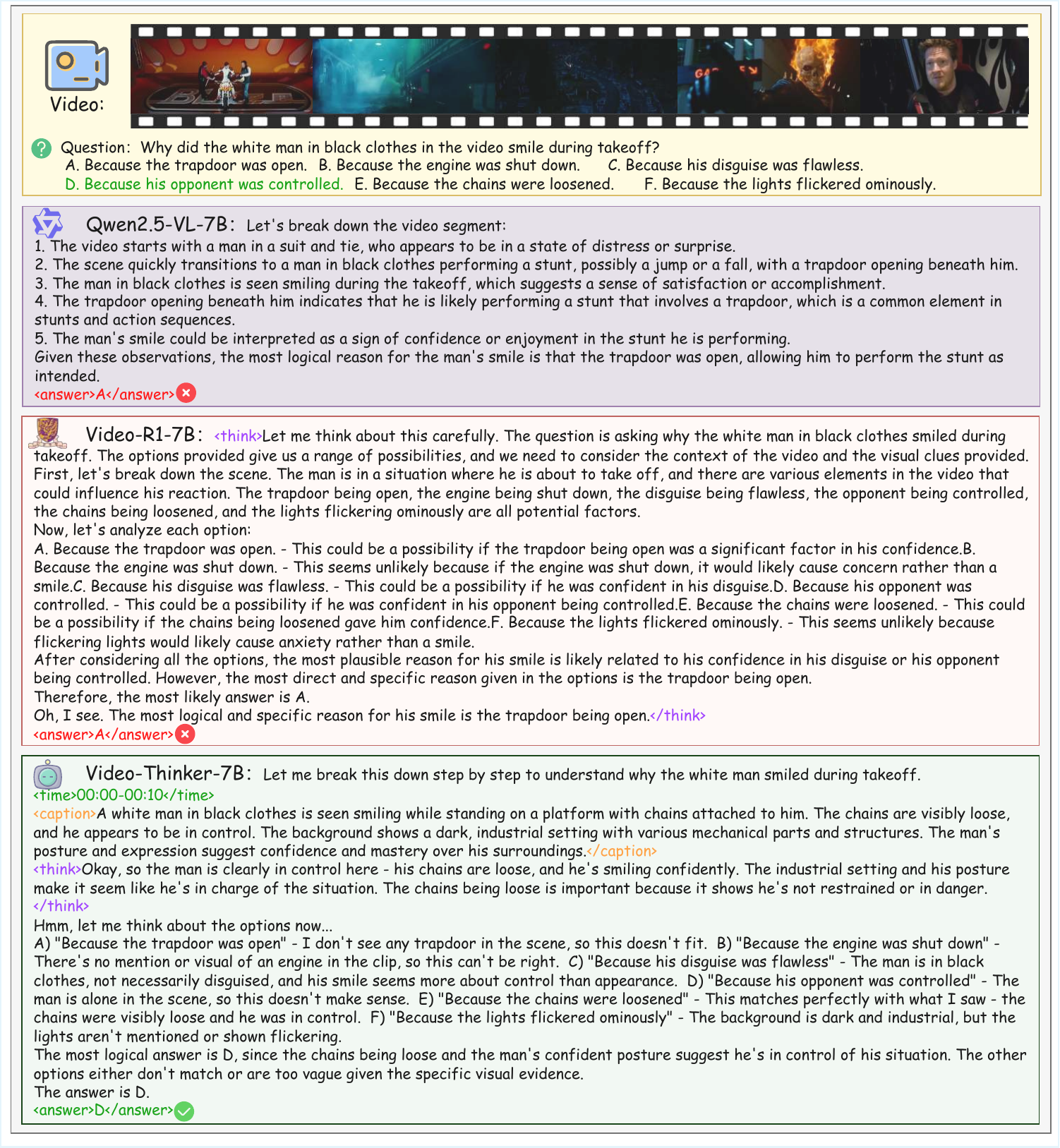}
    \vspace{-7mm}
    \caption{An example of Video-Thinker-7B's reasoning output on CG-Bench dataset}
    \label{fig:data-pipe6}
\end{figure*}

\begin{figure*}[h]
    \centering
    \includegraphics[width=\textwidth]{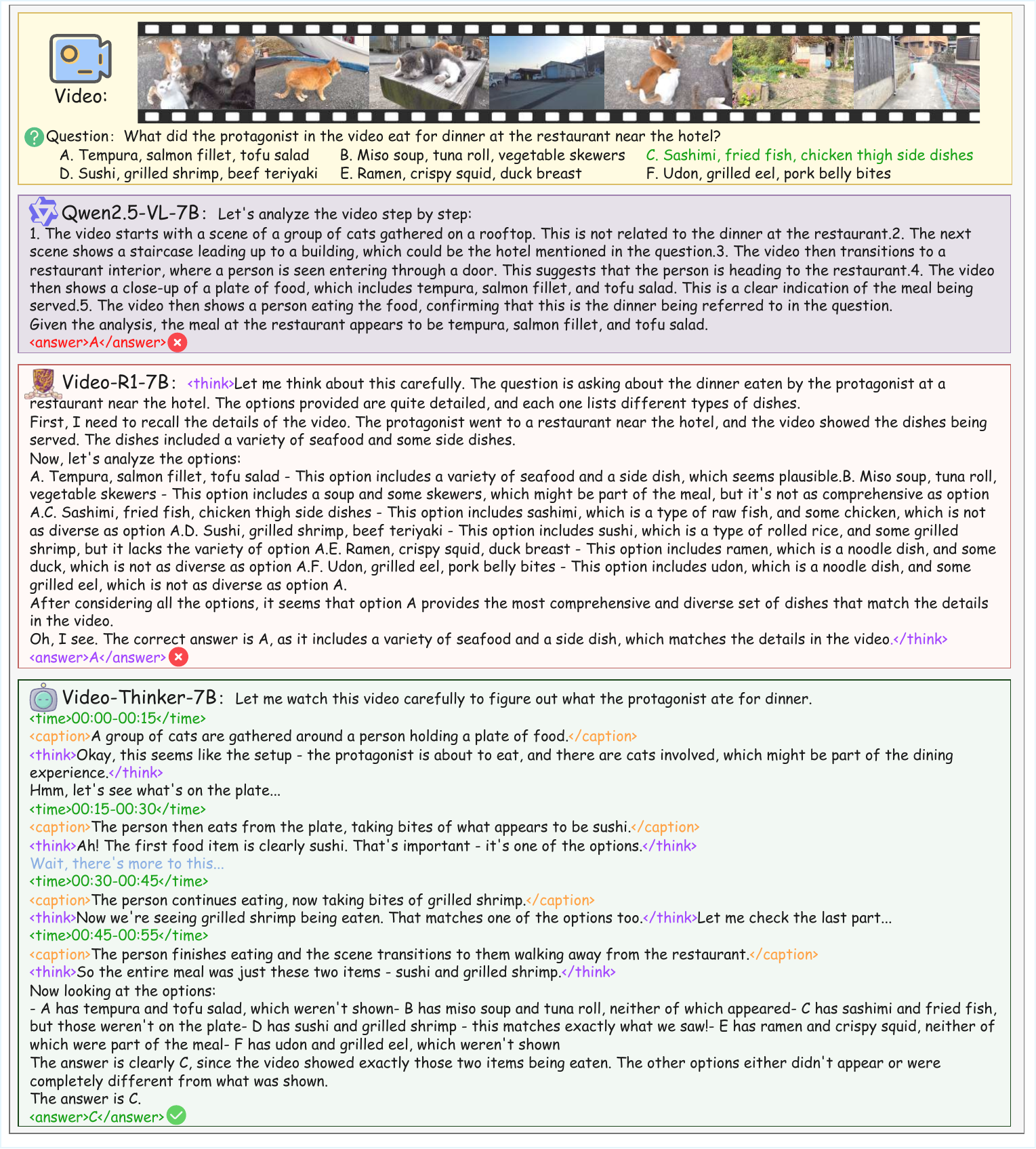}
    \vspace{-7mm}
    \caption{An example of Video-Thinker-7B's reasoning output on CG-Bench dataset}
    \label{fig:data-pipe7}
\end{figure*}

\begin{figure*}[h]
    \centering
    \includegraphics[width=\textwidth]{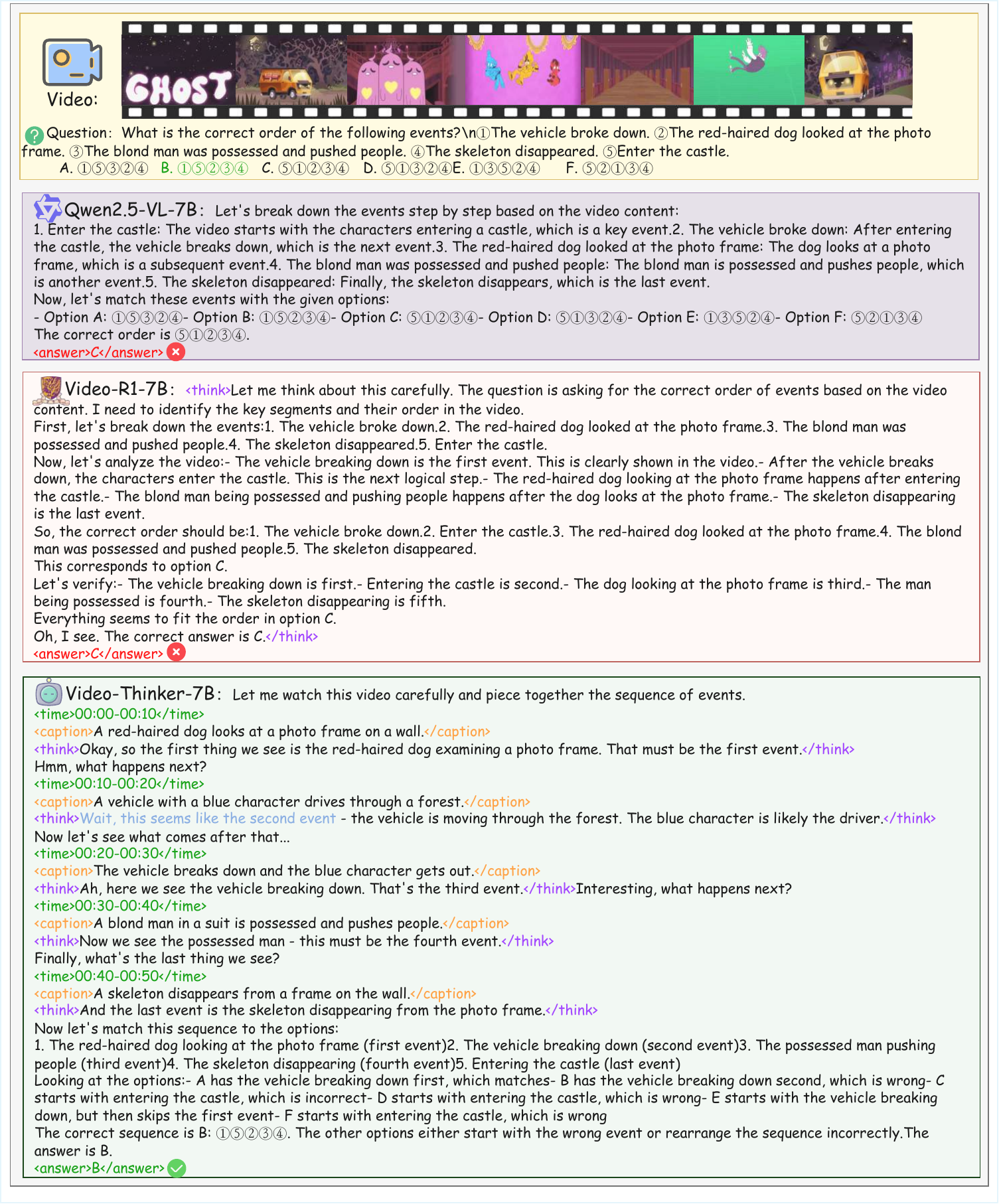}
    \vspace{-7mm}
    \caption{An example of Video-Thinker-7B's reasoning output on Video-Holmes dataset}
    \label{fig:data-pip8}
\end{figure*}

\begin{figure*}[h]
    \centering
    \includegraphics[width=\textwidth]{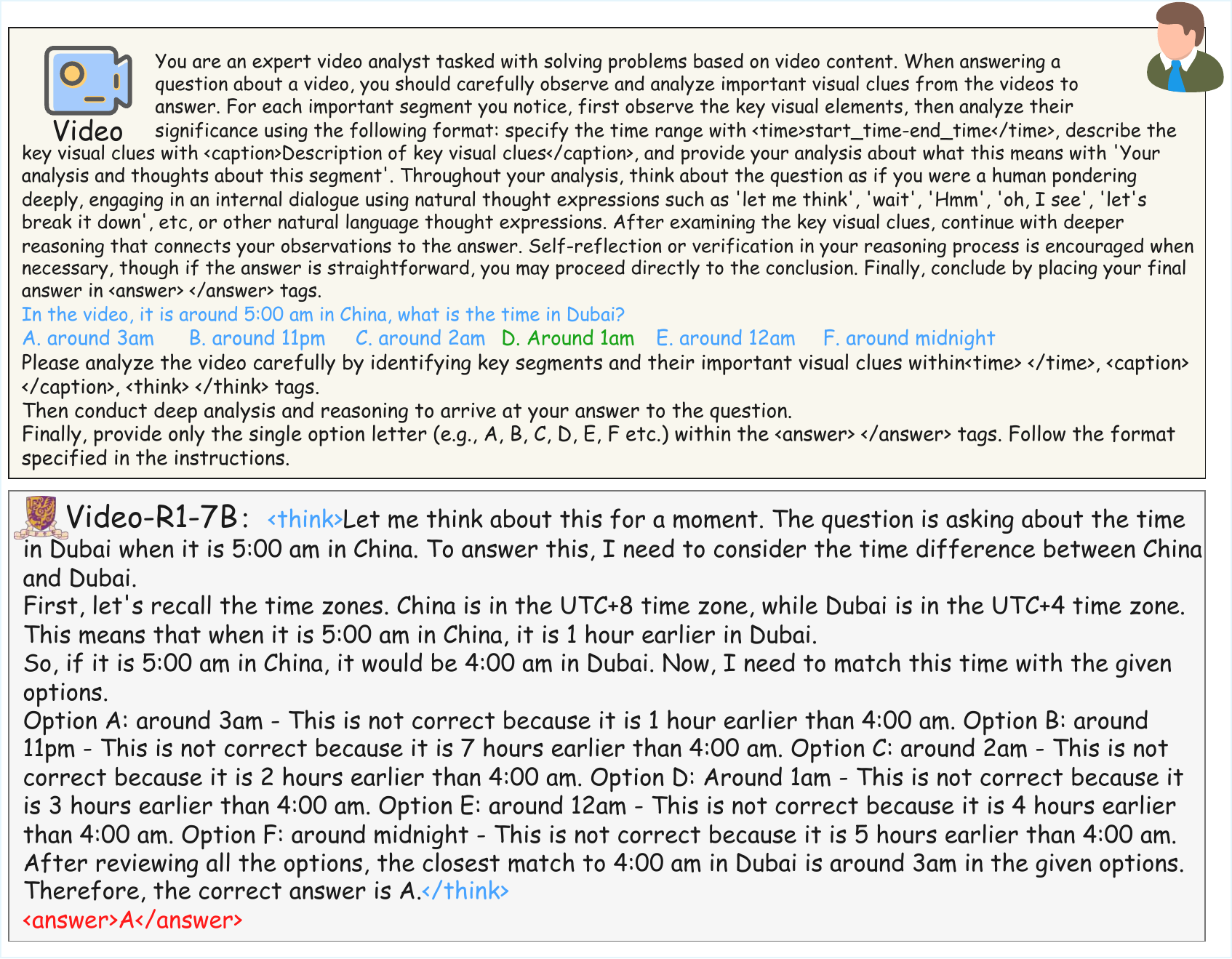}
    \vspace{-7mm}
    \caption{An example demonstrates Video-R1-7B's inability to follow instructions for generating temporal grounding content within \texttt{\green{<time></time>}} tags, thereby illustrating the rationale behind the statement in Section~\ref{sec:indepth}: ``Note that Video-R1 is excluded from this evaluation due to its inability to follow our prompt to generate temporal annotations within our templates.''.}
    \label{fig:data-pipe-not-follow}
\end{figure*}

\section{Cases}
\label{app:case}
In addition to the cases presented in Figure~\ref{fig:case}, we provide supplementary examples of Video-Thinker-7B's performance across diverse datasets in Figures~~\ref{fig:data-pipe1}, \ref{fig:data-pipe2}, \ref{fig:data-pipe3}, \ref{fig:data-pipe4}, \ref{fig:data-pipe5}, \ref{fig:data-pipe6}, \ref{fig:data-pipe7}, which demonstrate the model's capacity for iterative reasoning and error correction. This self-corrective behavior suggests that Video-Thinker transcends simple pattern matching and instead engages in a dynamic internal feedback mechanism.

\section{Use of LLMs}
During the preparation of this manuscript, we made limited use of publicly available large language models (LLMs) to assist with English writing. 
All technical content, including the formulation of ideas, design of methodologies, implementation of experiments, and interpretation of results, was entirely conceived and written by the authors without the involvement of LLMs. The role of LLMs was strictly confined to stylistic and linguistic improvements, in a manner comparable to grammar- or spell-checking software. 
We ensured that no novel research insights, data, or analyses were generated by LLMs, and all scientific claims and results presented in this work remain the sole responsibility of the authors.

\end{document}